\documentclass[10pt,twocolumn,letterpaper]{article}

\usepackage{wacv}              
\usepackage[accsupp]{axessibility}
\usepackage{multirow}
\usepackage{algorithm}
\usepackage{algpseudocode}

\DeclareMathOperator*{\argmax}{argmax}

\usepackage[table]{xcolor} 
\usepackage{colortbl}       
\usepackage{wrapfig,lipsum,booktabs}
\usepackage{graphicx}
\definecolor{wacvblue}{rgb}{0.21,0.49,0.74}
\definecolor{verylightgreen}{rgb}{0.21,0.49,0.74}

\usepackage[pagebackref,breaklinks,colorlinks,allcolors=wacvblue]{hyperref}

\def\wacvPaperID{1805} 
\def\confName{WACV}
\def\confYear{2026}

\title{LogicCBMs: Logic-Enhanced Concept-Based Learning}

\author{Deepika SN Vemuri$^{1}$ \quad Gautham Bellamkonda$^{1, 3}$ \quad Aditya Pola$^{1}$ \quad Vineeth N Balasubramanian$^{1, 2}$ 
\\
\normalsize{$^{1}$Indian Institute of Technology, Hyderabad, $^{2}$Microsoft Research, $^{3}$KLA} 
\\
\tt \small{\{ai22resch11001, ai24mtech02001, vineethnb\}@iith.ac.in}, gauthambellamkonda@gmail.com
}

\begin{document}
\maketitle
\begin{abstract}
Concept Bottleneck Models (CBMs) provide a basis for semantic abstractions within a neural network architecture. Such models have primarily been seen through the lens of interpretability so far, wherein they offer transparency by inferring predictions as a linear combination of semantic concepts. 
However, a linear combination is inherently limiting. So we propose the enhancement of concept-based learning models through propositional logic.
We introduce a logic module that is carefully designed to connect the learned concepts from CBMs through differentiable logic operations, such that our proposed LogicCBM can go beyond simple weighted combinations of concepts to leverage various logical operations to yield the final predictions, while maintaining end-to-end learnability. Composing concepts using a set of logic operators enables the model to capture inter-concept relations, while simultaneously improving the expressivity of the model in terms of logic operations.
Our empirical studies on well-known benchmarks and synthetic datasets demonstrate that these models have better accuracy, perform effective interventions and are highly interpretable.
\end{abstract}
    
\section{Introduction}
\label{sec:intro}
In recent times, building inherently interpretable models has gained prominence due to the drawbacks of post hoc explainability methods \cite{gradcam, shap, deeplift}. Concept-based models are a particularly promising direction \cite{cbms, oikarinen2023labelfree, hardcbms, sarkar2022cvpr,yuksekgonul2023posthoc}, where classes are considered to be composed of concepts that are interpretable units of human-understandable abstraction. For example, the model could learn to look for concepts like \textit{\{huge, gray, mammal\}} to classify an input as an \textit{elephant}. In these models, the concept-to-class mapping is often deliberately kept simple (usually just a linear layer) for interpretability, so that the importance of each concept can be directly inferred by examining the weights. However, such an approach may be restrictive and prevent the model from learning and leveraging higher-order relations between concepts.

\begin{figure}[!t]
  \begin{center}
\includegraphics[width=0.48\textwidth]{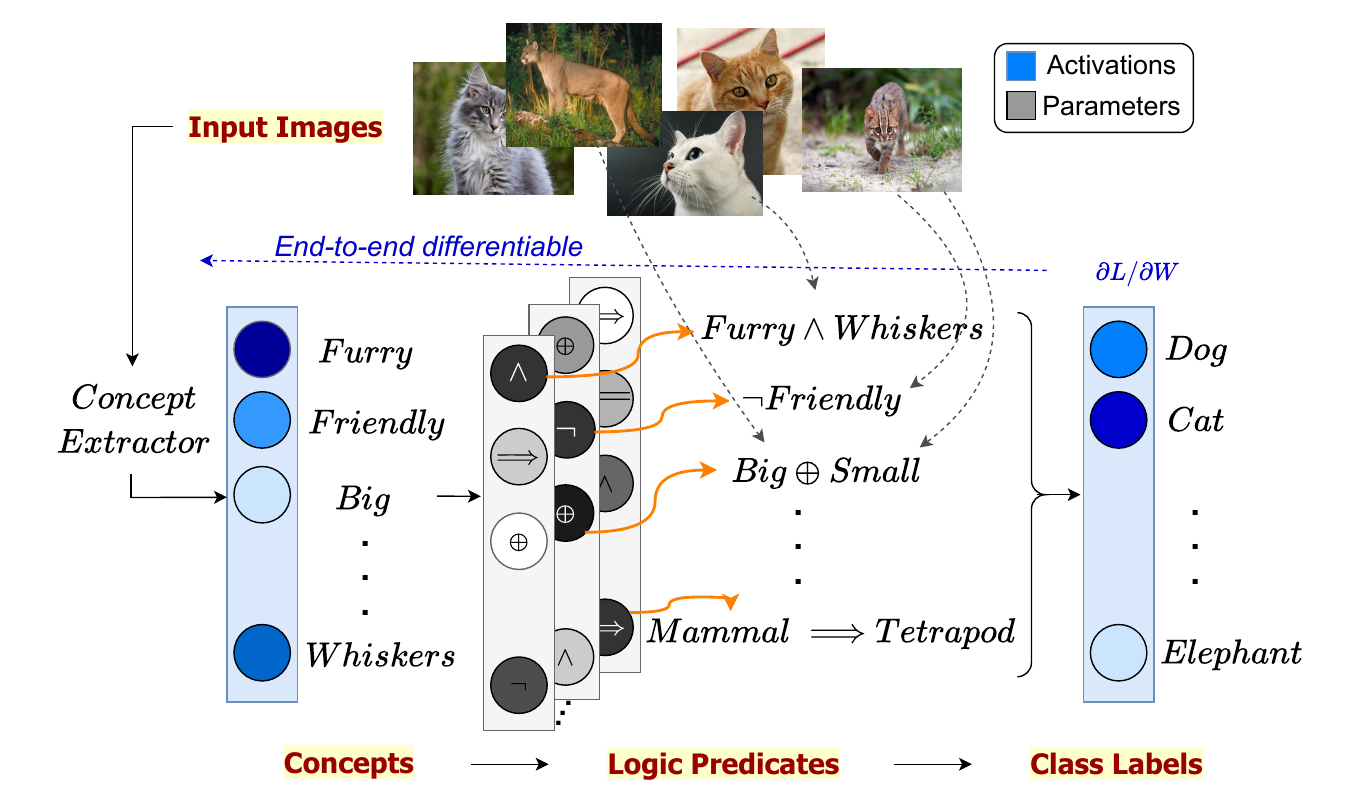}
\vspace{-12pt}
   \caption{\textbf{LogicCBMs: Overview of Our Approach.} We enhance concept-based learning models by including differentiable logic gates in the network. The model now forms logical compositions of concepts while predicting the class label for a given input image. (Dark shades indicate higher strength in the figure; e.g. \textit{Furry} is the strongest concept and \textit{Cat} is the predicted class.)} 
    \label{fig:teaser}
    \vspace{-15pt}
  \end{center}  
\end{figure}

\begin{figure}
  \begin{center}
\includegraphics[width=0.4\textwidth]{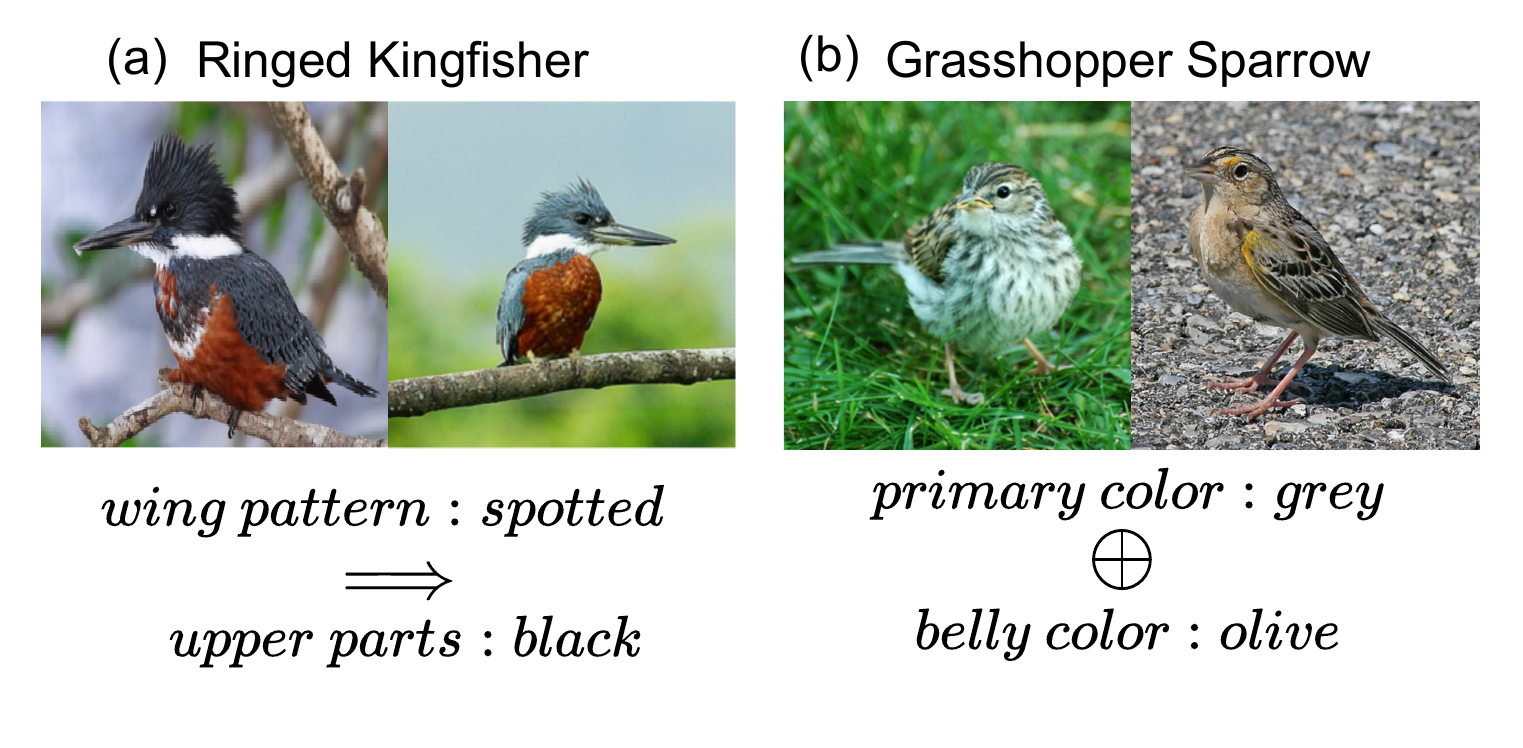}
\vspace{-12pt}
   \caption{\textbf{Logical predicates capture intra-class variability.} Examples of predicates (bottom) captured by our method for classes (top) in the CUB dataset (test set). (a) A \textit{Ringed Kingfisher} has black upper parts if it has a spotted wing pattern. This is captured by an IMPLIES operation. (b) A \textit{Grasshopper Sparrow} has its primary color as grey or belly color as olive (not both, not neither), captured by an XOR operation in our method.}
    \label{fig:cub_predicates}
    \vspace{-15pt}
  \end{center}  
\end{figure}

To illustrate this further, consider a model learning to recognize an \textit{arctic fox} class. Arctic foxes have either \textit{white fur} or \textit{brown fur} depending on the environmental conditions (equivalent to an exclusive OR operation). Evidently, a concept-based model will need to go beyond a linear layer to capture such nuances. We hence ask the question: \textit{while concept-based models are becoming increasingly popular, how do we go beyond the linear layer in the concept-class connections to capture richer relationships, while retaining interpretability?}

Logic operations present a natural choice to model concept-class relations in a structured manner (e.g. \textit{white fur} $\oplus$ \textit{brown fur} could indicate an \textit{arctic fox}). However, logic operations are non-differentiable by themselves, and are non-trivial to integrate inside neural network models in an end-to-end learnable manner. While there have been a few recent efforts on integrating logic into deep neural networks, they focus either on post-hoc analysis \cite{mu_compositions, rosa2023towards, Ciravegna_2023} or are focused on textual/tabular data \cite{lee2022selfexplaining}. We seek to address this need for integrating logic operations into concept-based models (CBMs) in an end-to-end learnable manner in this work. To this end, we propose LogicCBMs, a variant of CBMs where we introduce differentiable fuzzy logic gates that learn logic-based relationships between concepts and classes (as shown in \cref{fig:teaser}). LogicCBMs are end-to-end learnable, thus not compromising on classification performance, while allowing for rich interpretability of the concept-class relationships in terms of logic gates (see \cref{fig:cub_predicates}). Our studies show that logic gates are \textit{interpretable non-linearities}, and adding them in the network can be viewed as enabling the model with more capacity while retaining interpretability. 
Our method can be viewed as an initial effort that provides a pathway for integrating symbolic reasoning within concept-based learning architectures.
Our key contributions can be summarized as follows:
\begin{itemize}
    \item We introduce LogicCBMs, a technique to introduce logic operations into concept-based learning models in an end-to-end differentiable manner. The proposed approach builds logical compositions of concepts (predicates) and relates them to classes. 
    \item We provide a simple and efficient methodology to implement LogicCBMs that provides promising quantitative and qualitative improvements over existing concept-based models.
    \item We perform comprehensive experiments on well-known benchmark datasets for concept-based learning (CUB, CIFAR100 and AWA2) that demonstrate the promise of learning logic operations and show improved performance, not only in terms of quantative metrics but also qualitative interpretability analysis. Additionally, we also introduce a new worst-case analysis metric, Concept Correction Gain, to corroborate the usefulness of adding logic to concept-based architectures. As part of our empirical studies, we also introduce a simple synthetic dataset: \textit{CLEVR-Logic} to study logical relationships between concepts for validation of such methods in future.
    \item Our code and datasets will be made publicly available upon acceptance.
\end{itemize}

\section{Related Work}
\label{sec:relatedwork}
\vspace{-4pt}
\noindent \textbf{Concept-based Models.} Learning classes via concepts has been an actively growing area of research in recent years. Originally introduced in \cite{cbms}, later methods improved the method by addressing concept leakage \cite{Marconato2022GlanceNetsIL}, introducing a bypass concept layer \cite{sarkar2022cvpr}, including uncertainty quantification \cite{Kim2023ProbabilisticCB}, incorporating robustness \cite{Sinha2022UnderstandingAE}, improving interactivity \cite{interactivecbms}, and extracting concepts that are more amenable to composition \cite{SteinTowardsCI}. More recent efforts attempted the use of LLMs for concept guidance and expert annotations in \cite{oikarinen2023labelfree, LaBo}. Other efforts have included increasing model capacity by adding additional unsupervised concepts \cite{mahinpei2021promisespitfallsblackboxconcept, unsupervisedcbsm}, building concept bases for such models \cite{yuksekgonul2023posthoc} and making black-box models intervenable \cite{intervenable}. Inter-concept relations in these models have been studied from a concept representation space perspective \cite{raman2024understandinginterconceptrelationshipsconceptbased}, from  leveraging concept correlations \cite{stochasticcbms}, and from an energy-based modeling perspective \cite{xu2024energybased}. However, none of these efforts model the interactions between concepts that drive the predictive performance of a network, we present the integration of logic operations as a solution.

\vspace{1mm}
\noindent \textbf{Logic-based Explainability.} Concurrently, away from concept-based literature (focus of our work), there have been efforts to extract the underlying logic of a given model. Logic-explained networks \cite{Ciravegna_2023} operate on interpretable features, however, their logic explanations are obtained by building a truth table of binarized concepts from a fully trained model. SELOR \cite{lee2022selfexplaining} integrates logic into the model design via a probabilistic framework for logic rule generation; their work focuses however on tabular or textual data alone. \cite{mu_compositions, rosa2023towards} proposed post-hoc techniques for generating logical compositions of explanations. 
In contrast to these efforts, our proposed framework works learns the underlying logic structure among intermediate concepts emergent during training via differentiable logic on visual inputs.

\vspace{1mm}
\noindent \textbf{Differentiable Logic and Neurosymbolic Methods.} Efforts in integrating logic into neural network models is a nascent area, with most previous efforts focusing on neurosymbolic approaches for tasks like inductive logic programming \cite{sen2022neuro}, reinforcement learning \cite{zimmer2023differentiable} and abstract reasoning \cite{shindo2024learning}.
Some of these efforts pre-specify a set of rules, learning their probabilities differentiably \cite{manhaeve2018deepproblog, li2023scallop}, while a few others extract logic rules from examples by defining some kind of continuous relaxation over a discrete space such as the space of first-order logic (FOL) programs \cite{Zimmer2021DifferentiableLM}.
Fewer efforts attempt to use differentiable logic for better representation learning, viz. for better entity representation in knowledge graphs \cite{han2023logical} and for learning FOL rules for knowledge-based reasoning \cite{NIPS2017_0e55666a}.
On the other hand, our work focuses on integrating differentiable learnable logic into concept-based models, which allows a pathway to perform logic-based classification through concepts that capture intermediate latent semantics in a given model. Appendix \cref{tab:neurosym_comparison} provides a more comprehensive comparison.
\begin{figure*}[!t]
  \begin{center}
\includegraphics[width=460pt]{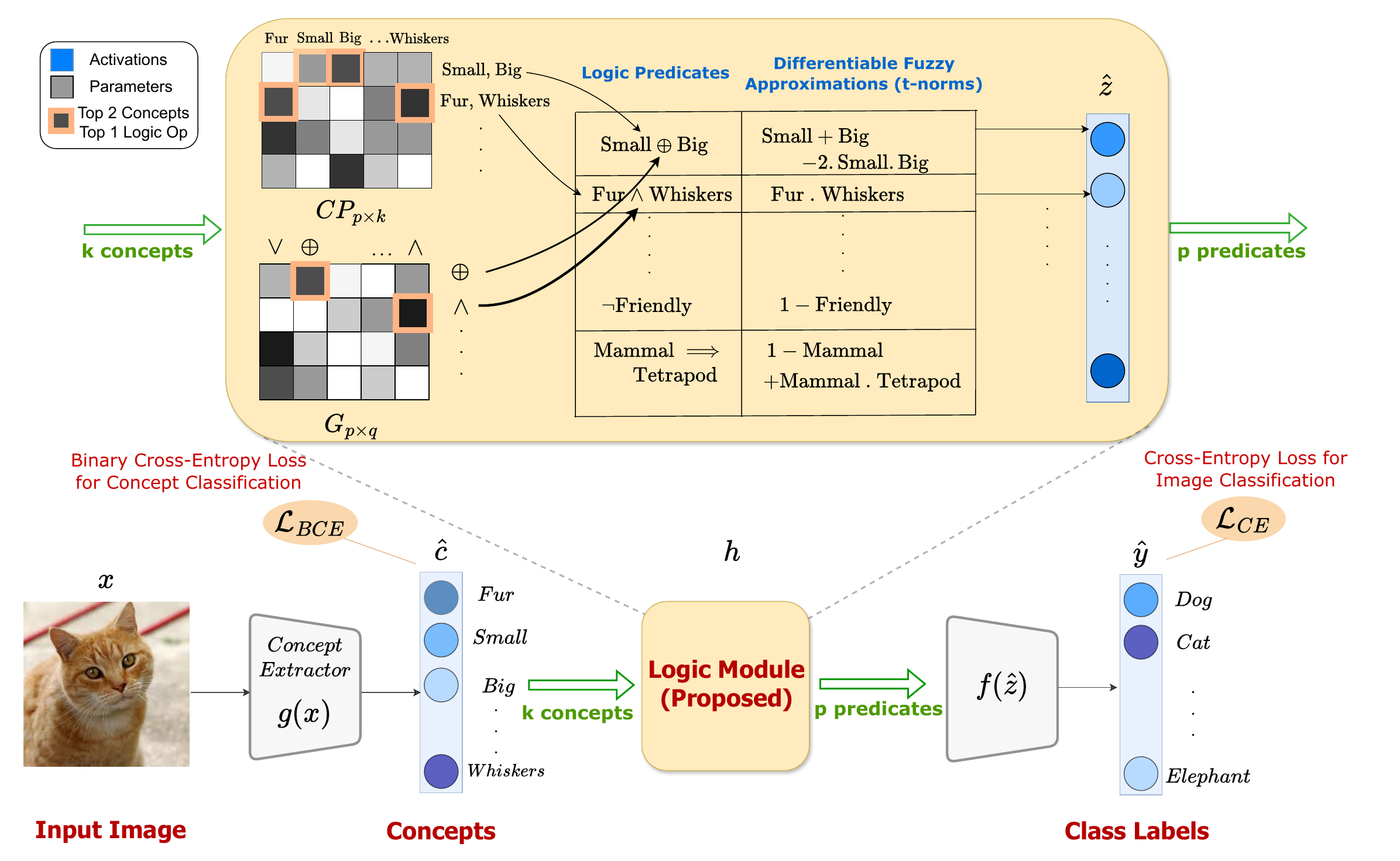}
\vspace{-6pt}
   \caption{\textbf{LogicCBM Architecture.} The proposed logic module/layer is added to a CBM after the concept layer $g(\cdot)$. To implement the logic module, we use two matrices: $CP$ (Concept Pairs) and $G$ (Logic Gates) after the concept layer to learn predicates using differentiable fuzzy logic operations. Our framework is end-to-end differentiable with a subsequent linear layer $f(\cdot)$ that learns the final predicate-class mapping to output the class label prediction.}
    \label{fig:arch_diagram}
    \vspace{-15pt}
  \end{center}  
\end{figure*}

\section{LogicCBMs: Methodology}
\label{sec:method}

\noindent \textbf{Preliminaries and Notation.} We follow the setup introduced by \cite{cbms} and define a concept-based model as a model that learns a mapping from $X \mapsto Y$ via an intermediate concept encoder $g(\cdot)$. These models learn from a three-tuple dataset $\mathcal\{X, C, Y\}$ where $X \in \mathbb{R}^{m}$, $C \in \mathbb{R}^{k}$, $Y \in \mathbb{R}^{n}$ and $m, k, n$ correspond to the dimensionalities of the image, concept and label spaces respectively.  Each prediction is of the form $\hat{y} = f(g(x))$ where $g \colon X \mapsto C$ (e.g. \textit{bird} image $\rightarrow \{\textit{white body},\, \textit{flat yellow bill},\, \dots,\, \textit{orange legs}\}$) is the concept encoder, and $f \colon C \mapsto Y$ (e.g. $\{\textit{white body},\, \textit{flat yellow bill},\, \dots,\, \textit{orange legs}\} \rightarrow \textit{Duck}$) is an interpretable predictor network..

\vspace{1.5mm}
\noindent \textbf{Conceptual Framework.} We define a \textit{predicate} to be a logical composition of concepts of the form $z_i$ = ($c_1$ $\texttt{op}_1$ $c_2$ $\texttt{op}_2$ $c_3$ ...), where $c_i \in C$ is a concept and  $\texttt{op}_j$ is a logic gate operation; for example, ($\textit{big} \oplus \textit{small}$), ($\textit{orange} \land \textit{vegetable} \land \textit{healthy}$). We are interested in learning the set of predicates logically entailed by each class:
\begin{equation} \label{eq:1}
 y_i \leftarrow w_1 . z_1 + w_2 . z_2 + w_3 . z_3 + ...
\end{equation}

\noindent For example, as shown in \cref{fig:teaser}, the \textit{Cat} class could assign a higher weight to predicates such as ($\textit{Furry} \land \textit{Whiskers}$), ($\textit{Mammal} \implies \textit{Tetrapod}$) and $\neg(\textit{Soft Nails})$.

\cref{fig:arch_diagram} shows the overall architecture of our approach. As shown in the figure, we do not make any changes in $g(\cdot)$, the backbone concept extractor, thus following a standard vanilla CBM in how it obtains concepts from input image samples. We introduce a logic module (which can consist of multiple logic gate layers) following the concept encoder to extract logical predicates, which are then sent to a classifier. We now write each prediction as: $\hat{y} = f(h(g(x)))$, where $h \colon \{0, 1\}^{k} \mapsto\mathbb{R}^p$ (\{\textit{white body}, \textit{flat yellow bill}, \textit{orange legs}\} $\rightarrow$ \{\textit{white body} $\land$ \textit{orange legs}, $\neg$ \textit{mammal}\}) and $f \colon \mathbb{R}^p \mapsto \mathcal{Y}$ (\{\textit{white body} $\land$ \textit{orange legs}\} $\rightarrow$ \textit{`Duck'}), where $p$ is the number of logic predicates.

\vspace{1.5mm}
\noindent \textbf{Learning from Differentiable Predicates. } For convenience of presentation, without loss of generality, we discuss the rest of our methodology for a logic module comprising one logic gate layer. Consider the logic gate layer to be composed of $p$ logic gate neurons, where each neuron is one of $q$ possible logic gates. 
There are two steps involved to learn predicates: (i) \textit{Concept pairing}, and (ii) \textit{Differentiable logic learning}. We describe each of these below.

\noindent (i) \textit{Concept Pairing}: We introduce two weight matrices: a \textit{concept pair matrix} $CP_{p \times k}$ and a \textit{logic gate matrix} $G_{p \times q}$. In this work, for convenience, we consider each logic gate neuron to take in one pair of concepts and thus form binary predicates. Stacking successive logic gate layers enables us to compose binary predicates to form more intricate $n$-ary predicates (which we show some initial results on later in this work, and is also an interesting future direction). As shown in \cref{fig:arch_diagram}, in order to determine these pairs, we extract the two concepts with the highest weight per row, which gives us a set of $p$ concept pairs.

(ii) \textit{Differentiable Logic Learning}: Logic gates are by themselves non-differentiable operations, and hence pose a constraint in incorporating them into end-to-end learnable architectures. 
In order to overcome this constraint, we take inspiration from fuzzy logic. In particular, we use t-norms, as in \cite{petersen2022deep, fuzzylogic}, which are fuzzy versions of logic gates to allow differentiability and backpropagation while passing the pairwise activations of the concept encoder through the logic layer. For example, if $z_1 = c_a \oplus c_b$ (XOR operation) and $c_a, c_b$ are concept activations, we can approximate this with the t-norm $c_a + c_b - 2 \cdot c_a c_b$. We use $q=16$ logic gate operations (as in \cite{petersen2022deep}) in our work herein (these gates are listed in \cref{tab:operators} in the Appendix). (While this list of gate operations is inherited from \cite{petersen2022deep}, our technical contributions lie in carefully designing and connecting the concept layer with the logic operations in a seamless learnable manner.) 

For each of the $p$ logic neurons, we compute all $q$ fuzzy logic operations on the concept pair assigned to it and learn a probability distribution to represent how important a logic gate is for that concept pair. The probability distribution for each logic neuron is captured by the $G$ matrix. The activation of a logic neuron then is the maximum of the weighted logic operation outputs, given by:
\vspace{-5pt}
\begin{equation}
    \hat{z_i} = \max_{i=1, \ldots, q} (g_i \cdot z_i(c_a, c_b)), \quad \sum_{i=1}^q g_i = 1, \, g_i \geq 0
\vspace{-3pt}
\end{equation}
\noindent where, $c_a$ and $c_b$ are the concept activations input to the $i^{th}$ logic neuron, $g_i$ is weight distribution and $\hat{z_i}$ is the activation of the $i^{th}$ logic neuron (from the $G$ matrix). Once the predicate activations are obtained, the model learns a predicate-class mapping using a linear layer.
\vspace{-5pt}
\begin{equation*}
    F_i = \sum_{j=1}^p V_{ij} \cdot \hat{z_j} \textnormal{ where, } 1 \leq i \leq n,
\vspace{-2pt}
\end{equation*}
\vspace{-3pt}
\begin{equation}
\hat{\mathbf{y}} = \sigma (F) \quad \textnormal{where}, F_i \textnormal{ is a logical formula}.
\end{equation}
\noindent where $\hat{z_j}$ is the $j^{th}$ logic neuron's activation. $V$ denotes the linear layer's parameters, and $\sigma$ is the softmax activation function. Each $F_i$ is a weighted sum of the learned predicates, which we refer to as a logic formula.

\vspace{1.5mm}
\noindent \textbf{Overall Training Procedure.} Our proposed architecture with the logic module does not entail any other loss terms, beyond the standard ones. This is in line with our objective to integrate logic learning seamlessly inside a concept-based learning model. Similar to existing efforts, the overall model is trained for concept classification using a binary cross-entropy loss ($\mathcal{L}_{BCE}$) (at the concept layer), and a final standard prediction cross-entropy loss ($\mathcal{L}_{CE}$) (at the output classification layer). Our overall loss thus remains similar to a CBM, given by:
\vspace{-5pt}
\begin{equation}
\mathcal{L} = \mathcal{L}_{CE} + \alpha \cdot \mathcal{L}_{BCE}
\vspace{-3pt}
\end{equation}
\noindent where $\alpha$ is a weighting hyperparameter. 

\section{Experimental Results and Analysis}
\label{sec:exps}

\textbf{Datasets: }We perform our experiments on a total of 
seven datasets, comprising standard natural benchmark datasets for CBMs (\textit{CUB200, AwA2, CIFAR100}), 
a large scale scene recognition dataset (\textit{SUN}) as well as synthetic datasets explicitly used to study the learned logic (\textit{XOR, 2XOR, CLEVR-Logic}).
Our natural datasets capture different levels of concept supervision: Caltech-UCSD Birds (CUB, \cite{cub_ds}), Animals with Attributes (AwA2, \cite{awa2}) and CIFAR100 \cite{Krizhevsky2009LearningML}. CUB and AwA2 have concepts for the classes annotated at an instance-level and class-level respectively, while for CIFAR100, we acquire class-level annotations from an LLM following the procedure outlined by \cite{oikarinen2023labelfree}. 
The SUN attribute dataset \cite{patterson2014sun} is also instance-level annotated.
More dataset details are in \cref{sec:dataset_details} in the Appendix.

\vspace{3pt}
\noindent \textbf{Baselines:}  We compare our approach with well-known concept-based learning models, in particular: (1) Vanilla \cite{cbms}, MLP and Boolean CBMs \cite{havasi2022addressing} (2) Label-Free CBMs \cite{oikarinen2023labelfree}, (3) Posthoc CBMs \cite{yuksekgonul2023posthoc}, (4) Sparse CBMs \cite{scbm} and (5) VLG CBMs \cite{vlg_cbms}.  Vanilla and Boolean CBMs use soft and hard concepts \cite{havasi2022addressing} in the training process of a CBM. An MLPCBM is a variant of a Vanilla CBM that replaces the linear concept-to-class mapping with a multilayer perceptron (MLP); one would expect this approach to capture richer concept-class relationships than a simple linear layer. Label-Free CBMs propose an LLM-based class-level concept annotation method, where they use CLIP-Dissect \cite{oikarinen2023clipdissect} for concept alignment. Posthoc CBMs propose a method for converting a blackbox model into a CBM using concept activation vectors \cite{cavs}. Sparse CBMs propose training CBMs using contrastive learning and self-supervision. Finally, VLG-CBMs use grounded open-domain object detectors to enhance the faithfulness of concept learning.

\definecolor{verylightgreen}{rgb}{0.9, 1.0, 0.9}
\begingroup
\setlength{\tabcolsep}{4pt} 
\hspace*{-0.5cm} 
\begin{table}[!t]
\centering
\begin{small}
\begin{tabular}{lccc} 
\hline \hline
\multicolumn{1}{c}{\textsc{Model}} & \textsc{CUB} & \textsc{AwA2} & \textsc{CIFAR100} \\
\hline
\textsc{Vanilla CBM} {\footnotesize \textcolor{gray}{\cite{cbms}}} & 75.20 \tiny{$\pm$ 0.79} & 88.81 \tiny{$\pm$ 0.52} & 55.39 \tiny{$\pm$ 0.62} \\
\textsc{MLP CBM} {\footnotesize \textcolor{gray}{\cite{havasi2022addressing}}} & 72.63 \tiny{$\pm$ 0.25} &  89.15\tiny{$\pm$0.23 } &  
65.00 \tiny{$\pm 0.31$} \\
\textsc{Boolean CBM} {\footnotesize \textcolor{gray}{\cite{hardcbms}}} & 63.57 \tiny{$\pm$ 0.27} & 82.97 \tiny{$\pm$ 0.33} & 47.40 \tiny{$\pm$ 0.82} \\
\textsc{LFCBM} {\footnotesize \textcolor{gray}{\cite{oikarinen2023labelfree}}} & 74.29 \tiny{$\pm$ 0.24} & 89.76 \tiny{$\pm$ 0.20} & 65.16 \tiny{$\pm$ 0.14} \\
\textsc{Posthoc CBM} {\footnotesize \textcolor{gray}{\cite{yuksekgonul2023posthoc}}} & 64.65 \tiny{$\pm$ 0.08} & 89.14 \tiny{$\pm$ 0.04} & 51.33 \tiny{$\pm$ 0.02} \\
\textsc{Sparse CBM} {\footnotesize \textcolor{gray}{\cite{scbm}}} &  
70.28 \tiny{$\pm$ 1.05} & 84.34\tiny{$\pm$ 0.12} &  61.68 \tiny{$\pm$ 1.00} \\
\textsc{VLG-CBM} {\footnotesize \textcolor{gray}{\cite{vlg_cbms}}} & 60.38
\tiny{$\pm$ 0.00} &  - & 65.73 \tiny{$\pm$ 0.00} \\
\rowcolor{verylightgreen} \textsc{LogicCBM (Ours)} & \textbf{81.13} \tiny{$\pm$ 0.42} & \textbf{90.04} \tiny{$\pm$ 0.05} & \textbf{68.46} \tiny{$\pm$ 0.45} \\
\hline \hline
\end{tabular}
\end{small}
\vspace{-4pt}
\caption{Validation accuracies (\%) on the CUB, AwA2, and CIFAR100 datasets averaged over 3 seeds. Results for VLG-CBMs were taken from their paper and don't report results on AwA2.}
\vspace{-5pt}
\label{tab:main_table}
\end{table}
\endgroup

\begin{figure}
    \centering
\includegraphics[width=0.95\columnwidth]{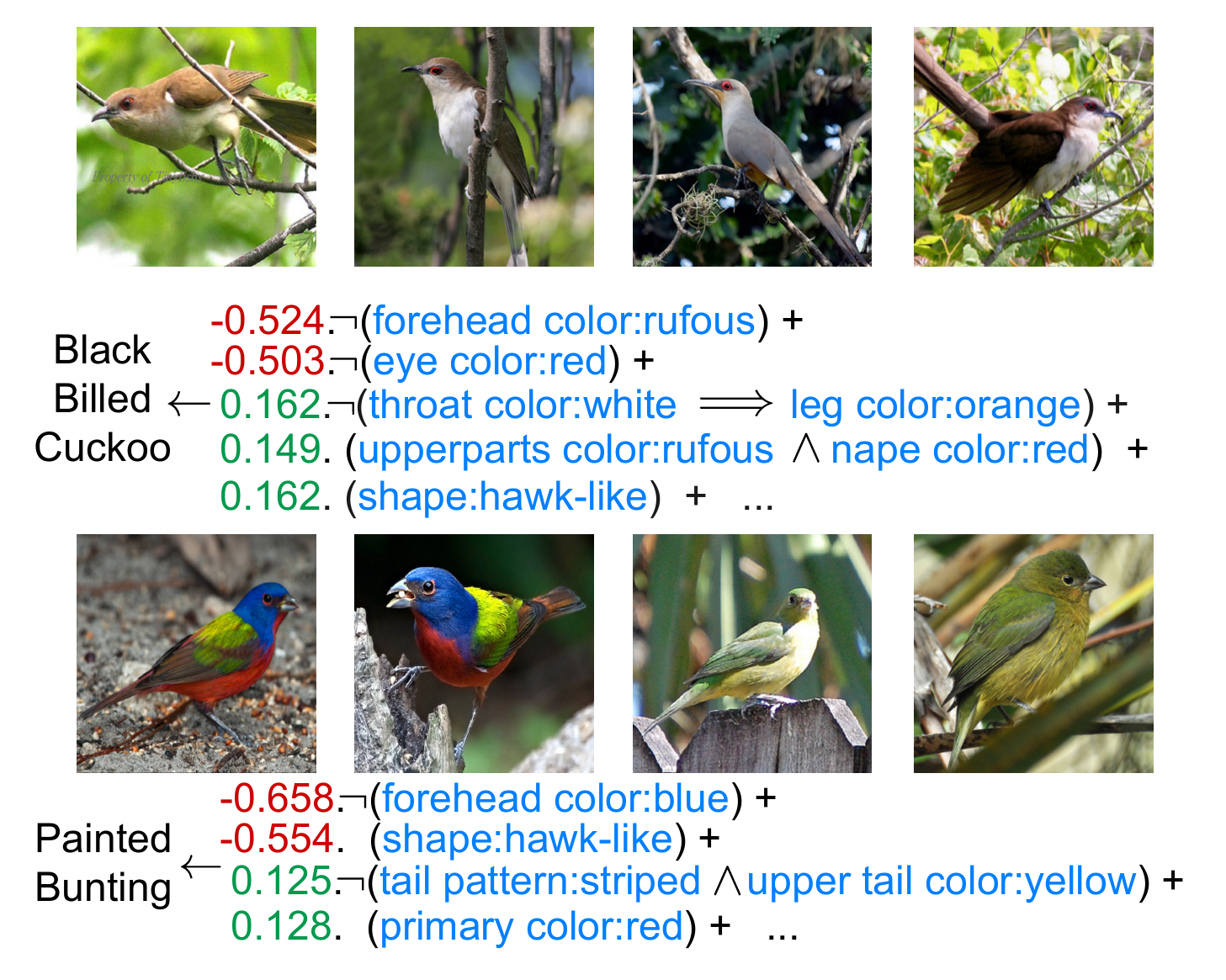}
\vspace{-8pt}
    \caption{Qualitative results showing examples of the class-level logic captured by LogicCBM on the CUB dataset}
    \label{fig:class_level_logic}
    \vspace{-13pt}
\end{figure}

\vspace{-6pt}
\noindent \textbf{Results:} 
\cref{tab:main_table} shows the experimental results on standard benchmark datasets. Our method provides an end-to-end training pipeline while outperforming prior approaches as shown in the table. We note that our implementation uses a single logic module/layer (we stick to binary predicates herein; extending to $n$-ary predicates would be an interesting future direction), and thus use consistently lesser parameters than a Vanilla CBM across datasets (please see \cref{tab:cbm_archs} in the Appendix). Our experiments with the concept pairing (CP) matrix also showed that both learned concept pairings and random concept pairings performed equivalently across datasets; we hence use the random pairing for its efficiency in practice. We hypothesize this works due to two key factors: (i) the subsequent learning process builds on this concept pairing to learn suitable predicate-class relationships; and (ii) there is sufficient redundancy in the logic module which allows appropriate logic relations to be learned. 
\cref{fig:class_level_logic} shows qualitative results of our method on the CUB dataset (more such results are included in \cref{sec:appendix-moreresults} in the Appendix). As observed from the figure, our logic module provides a concise, yet expressive set of interpretable units (predicates), which get the highest weight in our LogicCBM.

For completeness of this discussion, we also performed a focused study on comparing our LogicCBMs with a different family of recent methods that extract logic explanations from interpretable features -- LENs \cite{Ciravegna_2023}. Note that this method is not intended for end-to-end logic-based predictions (and hence is different from our core focus); however, for completeness of comparison, we train a $\psi$-net \cite{Ciravegna_2023} and a LogicCBM directly on concept ground truths (without any backbone network for concept extraction) from the CUB dataset. LogicCBMs outperformed the $\psi$-net (64\% vs 51\%), validating its usefulness for effectively learning concept-class relationships.

\subsection{Do these models learn meaningful logic?}
Since real-world datasets do not have explicit logic supervision, it is not straightforward to assess the correctness of the logic. We hence examine our method on synthetic datasets where the logic used to generate the data is known, allowing for validation of correctness of the model's learned logic. 

\vspace{1mm}
\noindent \textbf{Datasets, Baselines and Metrics:} We use three synthetic datasets: \textit{XOR} (proposed in \cite{espinosa2022concept}), \textit{2XOR} (a more complex version of XOR which we created that computes the XOR operation on three inputs) and \textit{CLEVR-Logic}, a new variant of CLEVR \cite{johnson2017clevr} which we generated to study logical relations among objects in images. In \textit{CLEVR-Logic} we define concepts to be a set of \textit{CLEVR} objects (sphere, cone, cube, cylinder) and specify classes as logical operations among these objects, which defines what is in the image (for example, \textit{sphere} $\oplus$ \textit{cone} could be one class which has images that exclusively contain either a \textit{sphere} or a \textit{cone}). CLEVR images are then generated per class following the class-specific logic. Sample images and their corresponding logical relationships are shown in \cref{fig:clevr_logic_images}. More details are included in \cref{sec:archnother_details} in the Appendix. All our code and datasets will be made publicly available upon acceptance.

\begin{figure}[t]
    \centering
\includegraphics[width=\linewidth]{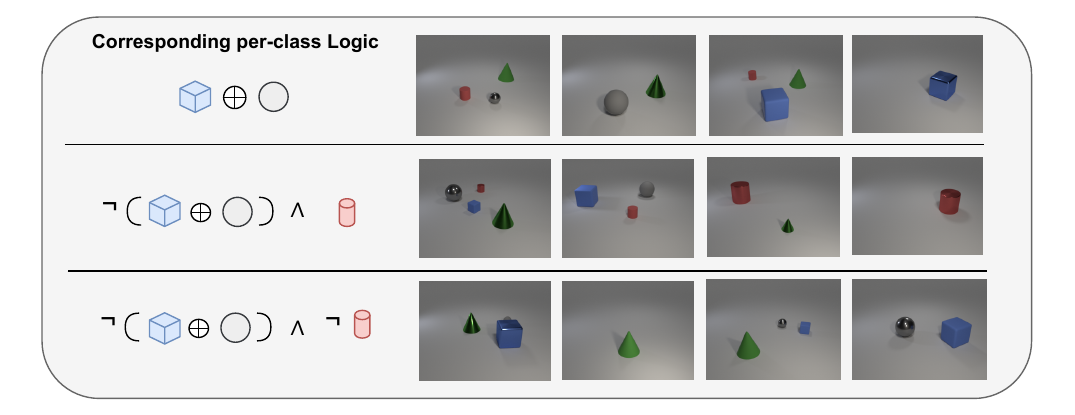}
    \vspace{-13pt}
    \caption{Sample images from our CLEVR-Logic dataset along with corresponding logic used to generate them.}
    \label{fig:clevr_logic_images}
    \vspace{-13pt}
\end{figure}

\begin{figure}
    \centering
    \includegraphics[width=220pt]{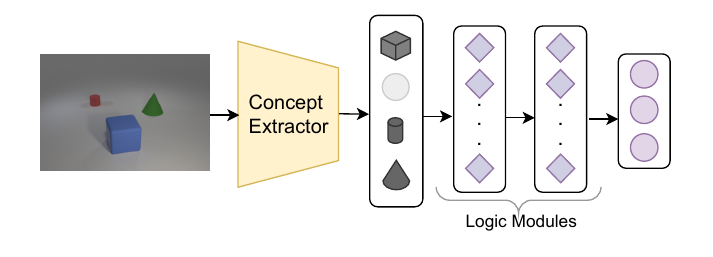}
    \vspace{-11pt}
    \caption{LogicCBM architecture for the CLEVR-Logic dataset. Polyhedra denote concepts, diamonds denote logic neurons, circles denote classes.}
    \label{fig:clevr_logic_arch}
\end{figure}

\begin{table}
\begin{small}
\begin{sc}
\centering
\resizebox{0.45\textwidth}{!}{
\begin{tabular}{c|c|c}
\hline \hline
\textbf{Dataset} & \textbf{DCR\cite{barbiero2023interpretable}} & \textbf{Our Method (LogicCBMs)} \\ \hline \hline
\multirow{2}{*}{\textit{XOR}} & ($c_1 \land$ $\sim c_2$) $\lor$ ($\sim c_1$ $\land c_2$) & $c_1 \oplus c_2$ \\ \cline{2-3}
& \# P : 180 & \# P : 16 \\ \hline \hline
\multirow{4}{*}{\textit{2XOR}} & ($\sim c_0 \land \sim c_1 \land c_2$)  & \multirow{3}{*}{$c_1 \oplus c_2 \oplus c_3$} \\
& $\lor$ ($\sim c_0 \land c_1 \land \sim c_2$) & \\
& $\lor \ldots$ & \\ \cline{2-3}
& \# P : 783 & \# P : 48 \\ \hline \hline
\end{tabular}
}
\vspace{-2pt}
\caption{Logic rules learned (along with number of parameters used: \#P) by DCR \cite{barbiero2023interpretable} and our method on \textit{XOR} and \textit{2XOR} datasets}
\vspace{-8pt}
\label{tab:dcr_vs_lcbm}
\end{sc}
\end{small}
\end{table}

\begin{table}[t]
\begin{small}
\begin{sc}
\centering
\resizebox{0.45\textwidth}{!}{
\begin{tabular}{|c|c|c|c|c|c|}
\hline \hline
\textbf{GT Rule} & \textbf{DCR\cite{barbiero2023interpretable} Error} & \textbf{LCBM Error} & \textbf{LCBM Rule} \\ \hline \hline
$c1 \oplus c2$ & 0.4 \% \footnotesize{\textcolor{gray}{$\pm$ 0.06}} & \textbf{0 \% \footnotesize{\textcolor{gray}{$\pm$ 0.00}}} & $c1 \oplus c2$ \\ \hline
$\neg(c1 \oplus c2) \land c3$ & 0.5 \% \footnotesize{\textcolor{gray}{$\pm$ 0.02}} & \textbf{0 \% \footnotesize{\textcolor{gray}{$\pm$ 0.00}}} & $\neg(c1 \oplus c2) \land c3$ \\ \hline
$\neg(c1 \oplus c2) \land \neg c3$ & 1.7 \% \footnotesize{\textcolor{gray}{$\pm$ 1.5}} & \textbf{0 \% \footnotesize{\textcolor{gray}{$\pm$ 0.00}}} & $\neg(c1 \oplus c2) \land \neg c3$ \\ \hline \hline
\end{tabular}
}
\vspace{-4pt}
\caption{Error rate results for DCR and our method (LCBM = LogicCBM) on \textit{CLEVR-Logic} dataset (all models trained on 2 seeds). We also show the ground truth logic and logic learned by our model, which matches the ground truth on all these experiments.}
\label{tab:clevr_results}
\vspace{-9pt}
\end{sc}
\end{small}
\end{table}

\vspace{1mm}
\noindent \textbf{Implementation Details:} For the \textit{XOR} and \textit{2XOR} datasets, our objective in the training process is to learn the correct logic predicate among the options (XOR is one of the possible logic gates in our logic module). To this end, we train single-layer and two-layer LogicCBM models for the two datasets respectively. For \textit{CLEVR-Logic}, we 
train a concept encoder to capture various CLEVR objects as concepts.
The extracted concepts are then passed through two logic modules (since some ground truth logic requires 3-ary predicates, as shown in \cref{fig:clevr_logic_images}) containing 15 logic neurons each. \cref{fig:clevr_logic_arch} shows our model architecture. 
The predicates themselves are the class labels here.
We also train DCR \cite{barbiero2023interpretable} models as a baseline on all three datasets for comparison. 

\vspace{1mm}
\noindent \textbf{Results:} \cref{tab:dcr_vs_lcbm} presents the results for the \textit{XOR} and \textit{2XOR} datasets. As evident in the table, our models learn more succinct and expressive logic, while also using significantly lesser parameters. The DCR models generate explanations only in terms of AND, OR and NOT, necessitating lengthy explanations, especially evident in the case of \textit{2XOR}. \cref{tab:clevr_results} shows the results (including the ground truth and learned predicates) for multiple runs on the \textit{CLEVR-Logic} dataset. Our LogicCBM model is reliably able to learn the underlying class-level logic. The DCR model misclassifies some input samples, as seen in the error rates shown in the table.

\subsection{More Analysis}


\noindent \textbf{4.2.1. Logic leads to better concept alignment} \\
Concept activations of samples belonging to the same class are expected to be close to each other. We capture this aspect of a model's behavior using \textit{concept alignment}, where we measure the average degree of similarity of concept activations among samples belonging to the same class. Let $n_i$ denote the number of samples belonging to class $i$ and $g$ indicate the model's concept encoder. We compute concept alignment using cosine similarity as below:
\vspace{-6pt}
\begin{equation}
    CA(i) = \sum_{j_1=1}^{n_i} \sum_{j_2=j_1+1}^{n_i} \cos(g(x_{j_1}),  g(x_{j_2}))
    \vspace{-2pt}
\end{equation}
\noindent We compare Vanilla and LogicCBMs, the results of which are shown in \cref{tab:cas_scores}. Our logic-based models have a consistently higher concept alignment across datasets. We provide some class-level analysis of this in the Appendix (\cref{sec:appendix-moreresults}).

\begin{table}[h]
\begin{sc}
\centering
\begin{footnotesize}
\begin{tabular}{|c|c|c|c|}
\hline \hline
\textbf{Method} & \textbf{CUB} & \textbf{AwA2} & \textbf{CIFAR100} \\
\hline \hline
Vanilla CBM & 0.8619 \tiny{$\pm$ 0.001} & 0.9754 \tiny{$\pm$ 0.021} & 0.9936 \tiny{$\pm$ 0.007} \\
\rowcolor{verylightgreen} LogicCBM & \textbf{0.9284} \tiny{$\pm$ 0.001} & \textbf{0.9810} \tiny{$\pm$ 0.026} & \textbf{0.9996} \tiny{$\pm$ 0.000} \\
\hline \hline
\end{tabular}
\vspace{-4pt}
\caption{Mean concept alignment scores on Vanilla CBM and LogicCBM. The higher the better.}
\label{tab:cas_scores}
\vspace{-3pt}
\end{footnotesize}
\end{sc}
\end{table}


\noindent \textbf{4.2.2. Logic can be used for finetuning} \\
We studied the possibility of using our logic module to finetune a pre-trained concept-based model. In certain use cases, it is possible that we may not want to train a LogicCBM from scratch (or architecturally change an existing concept-based model), but may be interested in enhancing an existing model's performance with additional logic. 
\setlength{\intextsep}{4pt}%
\setlength{\columnsep}{10pt}%
\begin{wrapfigure}[10]{r}{0.3\textwidth} 
  \centering
\includegraphics[width=0.35\textwidth, clip, trim=5 5 5 5]{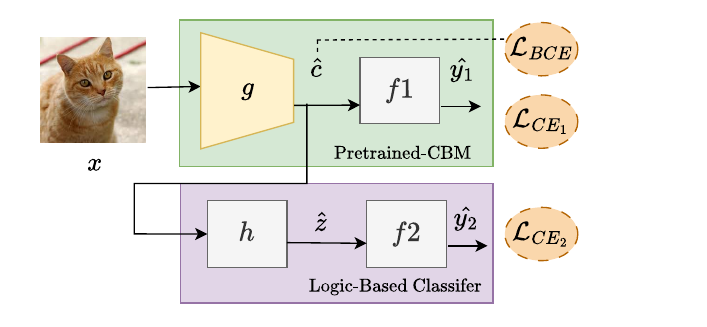}
\vspace{-12pt}
   \caption{Proposed architecture to use logic for finetuning an existing CBM.}
    \label{fig:feedback_arch}
\end{wrapfigure}
To this end, as shown in \cref{fig:feedback_arch}, we propose the use of our logic module in a separate logic classification head in the architecture ($f_2$), which takes in concept activations during finetuning. The baseline CBM's backbone can now leverage the logic head's gradient as well. The training objective in this case hence is given by:
\vspace{-2pt}
\begin{equation}
\mathcal{L} = \mathcal{L}_{CE_1} + \alpha \mathcal{L}_{CE_2} + \beta \cdot \mathcal{L}_{BCE}    
\vspace{-2pt}
\end{equation}
\noindent where a cross-entropy loss is used over both the heads (original and logic head), where $\alpha$ and $\beta$ are weighting hyperparameters. Our results on Vanilla CBMs using this approach are reported in \cref{tab:logic_finetuning} and show consistent gains and promise in such an approach. 
\begin{table}[h]
\begin{sc}
\centering
\begin{footnotesize}
\begin{tabular}{|c|c|c|c|}
\hline \hline
\textbf{Model} & \textbf{CUB} & \textbf{AwA2} & \textbf{CIFAR100} \\
\hline \hline
Vanilla CBM & 75.20 \tiny{$\pm$ 0.79} & 88.81 \tiny{$\pm$ 0.52} & 55.39 \tiny{$\pm$ 0.62} \\
Vanilla CBM + L & 79.21 \tiny{$\pm$ 0.08} & 89.56 \tiny{$\pm$ 0.07} & 65.89 \tiny{$\pm$ 0.72} \\
\hline \hline
\end{tabular}
\vspace{-3pt}
\caption{Classification accuracy of CBMs obtained by finetuning using our logic module (LF: Logic for Finetuning).}
\label{tab:logic_finetuning}
\vspace{-3pt}
\end{footnotesize}
\end{sc}
\end{table}

\vspace{0.5mm}
\noindent \textbf{4.2.3. Logic makes interventions more effective} \\
One way to study the correctness of the learned logic using LogicCBMs is to intervene on concepts in test samples, and observe the outcomes. In order to study this, we perform interventions on misclassified samples at test time, where we randomly choose $k$ concepts (output of concept layer) obtained from the respective data sample and replace them with their ground truths. On all baseline models, we perform $k \in \{4, 8, 10\}$ such test-time interventions; and on our LogicCBM model, we perform k/2 predicate interventions (i.e. $\{2, 4, 5\}$) for fairness of comparison, as each logic predicate is composed of two concepts. 
\setlength{\intextsep}{4pt}%
\setlength{\columnsep}{8pt}%
\begin{wrapfigure}[14]{r}{0.27\textwidth} 
\centering
    \includegraphics[width=\linewidth]{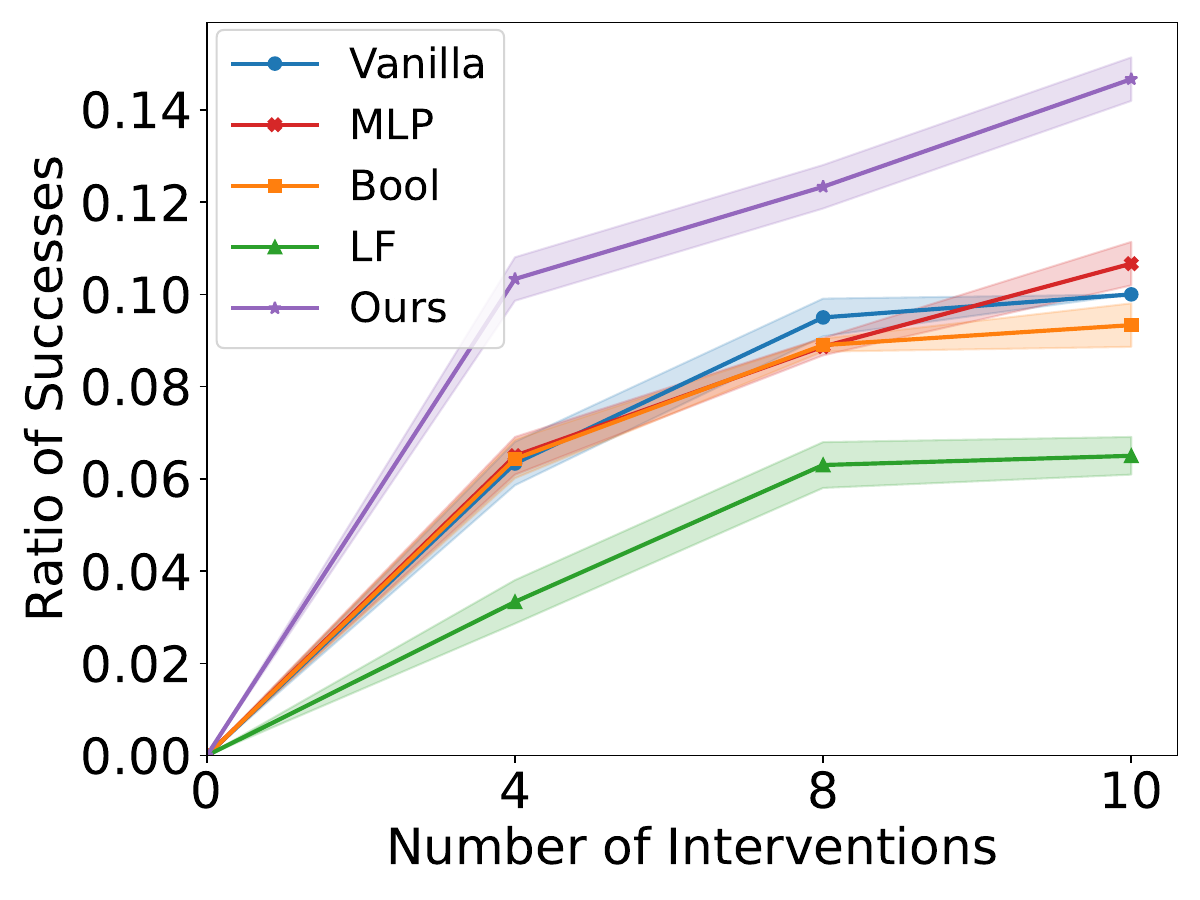}
    \vspace{-20pt}
    \caption{Ratio of test-time intervention successes on baselines vs our CBM method on the CUB dataset. $x$-axis indicates number of interventions performed.}
    \label{fig:cub_interventions}
    \vspace{-13pt}
\end{wrapfigure}
Also, as logic predicates do not have direct ground truths available, we manually replace the predicate in the misclassified sample with the expected logic neuron operation on the corresponding ground truth concepts. We perform this evaluation on CUB, as it is relatively the most difficult dataset.
\cref{fig:cub_interventions} shows the 
the ratio of intervention successes as the fraction of the number of misclassified samples that change to the correct prediction (after this intervention), among all misclassified samples for a given method.
Ideally, a higher ratio would indicate a better concept-class relationship learned. As the plot shows, we see that LogicCBMs have the most effective successful interventions. Some baseline methods (Posthoc and Sparse CBMs) don't have concept ground truths at a sample/class level, and hence could not be reported herein. 


\vspace{1mm}
\noindent \textbf{4.2.4. The diversity of logic gates used matters}\\
As stated earlier, the number of logic gates used in LogicCBMs are $q = 16$ (as listed in \cref{tab:operators} in the Appendix). \cref{fig:gate_barplots} shows the distribution across logic gates learned by the models on the CUB and AwA2 datasets (plots for CIFAR100 is in Appendix \cref{sec:appendix-moreresults}). The bar plots show the use of all gates in each of the datasets. Some gates like NOT are relatively less used on some of the datasets, possibly because the other predicates had a stronger impact on the prediction. 
\begin{figure}[h]
    \centering
    \includegraphics[width=\linewidth]{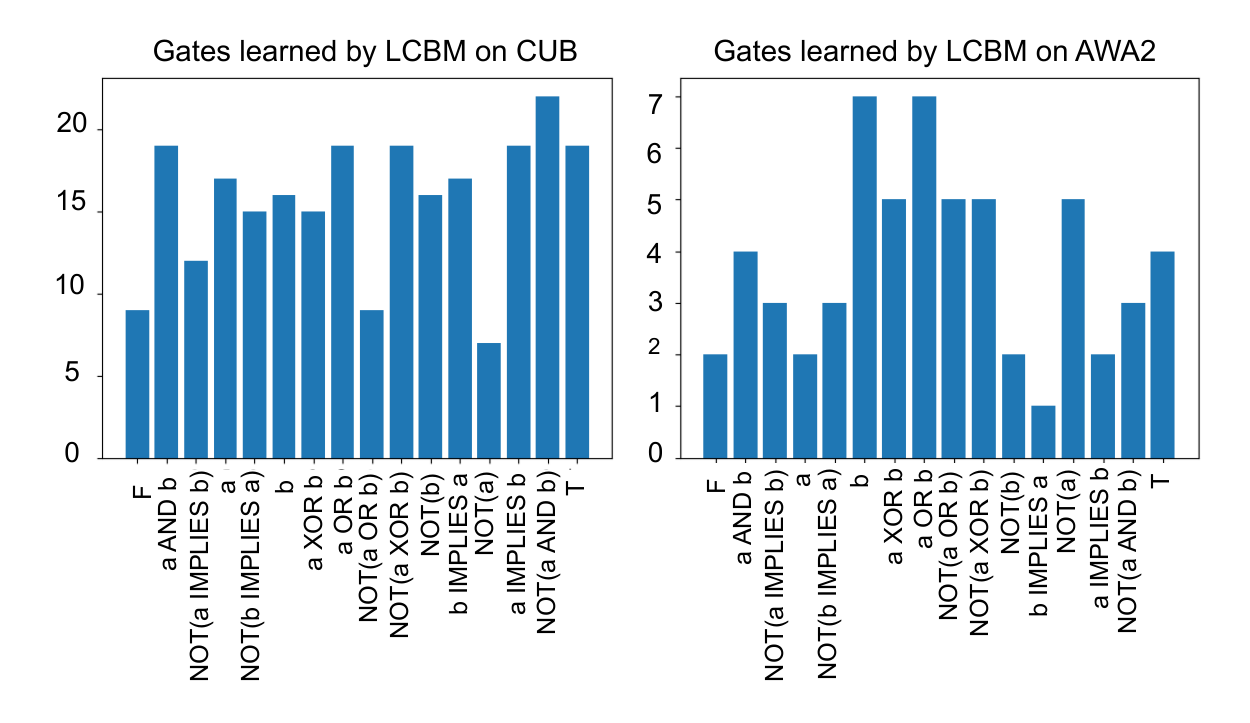}
    \vspace{-20pt}
    \caption{The distribution of logic gates learned by our LogicCBM models on the CUB and AwA2 datasets. 
    See Appendix \cref{sec:appendix-moreresults} for results on CIFAR100.}
    \label{fig:gate_barplots}
\end{figure}
To study this further, we reduce the number of logic gate types used to 8, thus reducing the diversity of possible logic operations, and study the accuracy on each of these datasets. In particular, the set of 16 logic gates is pruned to 8 by retaining only the simpler operations ($\land, \lor, 1, 0, c_1, c_2, \neg c_1, \neg c_2$) and removing the more expressive ones. \cref{tab:reduced_q} shows the results of this study, where we observe that the reduced number of logic gate types used has a marked impact on performance with a considerable drop in accuracy across datasets, particularly evident on CUB which we attribute to its fine-grained nature. This highlights the use of a diverse set of logic gates for better performance. Exploring an optimal set of logic gates for a given task can be another interesting direction of future work.
\setlength{\intextsep}{2pt}%
\setlength{\columnsep}{8pt}%
\begin{wrapfigure}[26]{r}{0.25\textwidth} 
    \centering
\includegraphics[width=\linewidth]{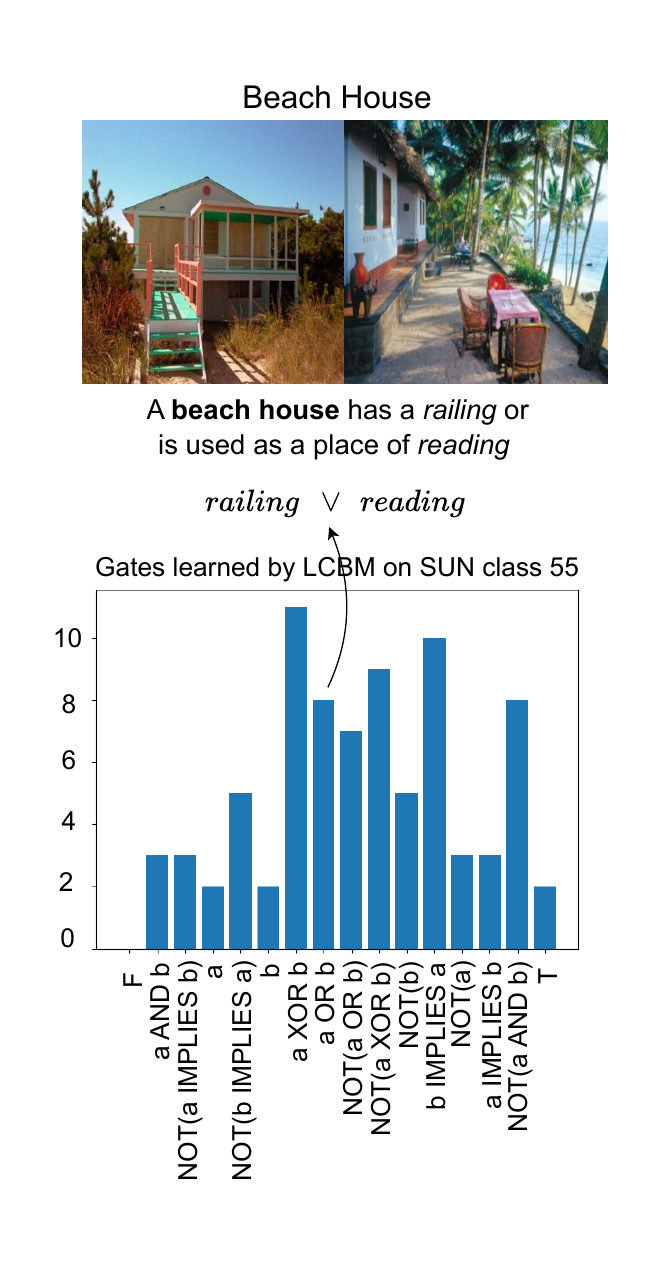}
    \vspace{-28pt}
    \caption{An example predicate learned by a LogicCBM on the SUN Attribute dataset for the \textit{Beach House} class. The bar plot indicates the distribution of logic gates learned for the class.}
    \label{fig:sun_result}
\vspace{-13pt}
\end{wrapfigure}

\begin{table}[t]
\centering
\begin{footnotesize}
\begin{sc}
\begin{tabular}{|c|c|c|c|}
\hline \hline
\textbf{\#Logic Types} & \textbf{CUB} & \textbf{AwA2} & \textbf{CIFAR100} \\ \hline \hline
16 & \textbf{81.13} $\pm$ \tiny{0.42} & \textbf{90.04} $\pm$ \tiny{0.05} & \textbf{68.46} $\pm$ \tiny{0.45} \\ \hline
8 & 
63.32 $\pm$ \tiny{0.62} & 86.39 $\pm$ \tiny{0.61} & 59.45 $\pm$ \tiny{0.11} \\ \hline \hline
\end{tabular}
\vspace{-4pt}
\caption{Drop in accuracy on reducing $q$ (number of logic operations considered) from 16 to 8. Results indicate the importance of diversity of logic gates used.}
\vspace{-20pt}
\label{tab:reduced_q}
\end{sc}
\end{footnotesize}
\end{table}

\begin{figure*}[t]
    \centering
    \includegraphics[width=464pt]{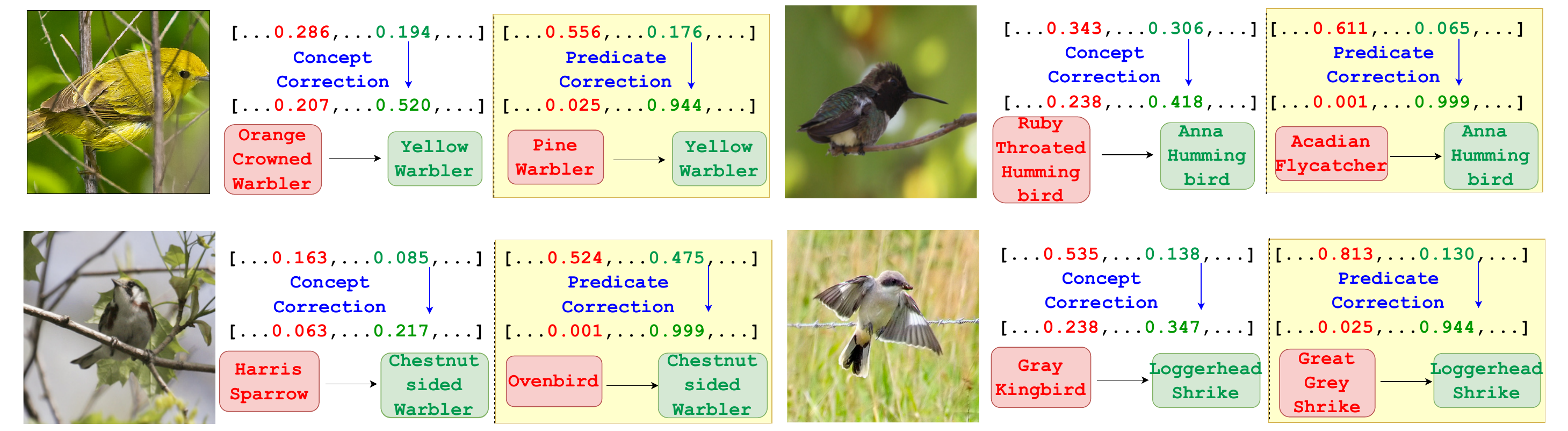}
    \vspace{-6pt}
    \caption{Some examples of pre- and post-correction confidences of a Vanilla CBM (left) and a LogicCBM (right, highlighted in yellow). Note the significantly improved confidences of the LogicCBM model post correction. }
    \label{fig:ccg_qualitative}
    \vspace{-6pt}
\end{figure*}

\noindent \textbf{4.2.5. Scaling to 500+ classes: LogicCBMs for scene recognition} \\
In order to study how LogicCBMs scale, we conduct experiments on the SUN attribute dataset \cite{patterson2014sun}, a large-scale scene recognition dataset with 500+ categories. LogicCBMs match the performance of a Vanilla CBM ($\approx$ 85\% validation accuracy on both models).
Importantly, we observed that the qualitative results showed improved performance; an example of the logic captured by our model is shown in \cref{fig:sun_result} along with the learned class-level logic gate distribution (more such results are in Appendix \cref{sec:appendix-moreresults}). This is also reflected in our other metric in Appendix \cref{tab:ccg_appendix}. We obtain the class-level logic distribution by running inference on all samples belonging to a chosen class and averaging the maximum activating logic gate for each logic neuron over all samples.

\subsection{CCG: A Worst-Case Metric for CBMs}
\vspace{-3pt}
Concept-based interpretable models are especially of importance in high-risk scenarios where there may be a high cost associated with an erroneous decision. We hence propose a metric to study CBM-related models in such a worst-case setting to assess the quality of concept/predicate-class relationships they have learned. 
Our evaluation is based on test-time corrections of concepts (or predicates). For each sample, we identify the most misleading concept (or predicate) and turn it off (or on, depending on ground truth for a given example). We then measure the resulting change in the model's prediction confidence, where confidence is measured as the softmax output of the ground truth class.
For example, certain species of dogs have pointy ears, although it is more common in cats. If a model incorrectly predicts \textit{'cat'} for such an image, we can test whether turning off the misleading concept \textit{pointy ears} increases the model’s confidence in the correct class (\textit{'dog'}).
A significant positive change in model confidence represents correctness of the learned concept-class relationships. We define this metric as \textit{Concept Correction Gain} (CCG). Ideally, the CCG would be high, indicating sensitivity to the right semantic cues.
CCG is similar in principle to activation patching \cite{activationpatching}. While activation patching is a mechanistic interpretability technique to analyze the behavior of a model's components, CCG quantifies a model's confidence when some misleading information is removed. 
Formally:
\begin{equation}
    \mathrm{CCG} = \frac{1}{|\mathcal{C}|} \sum_{i \in \mathcal{C}} \left[ \sigma\big(\check{y}_i\big)_{y_i^{\ast}} - \sigma\big(\hat{y}_i\big)_{y_i^{\ast}} \right]
\end{equation}
\noindent where, $y_i^\ast$ is the ground truth class from the $i^{th}$ sample, $\check{y}_i$ and $\hat{{y}}_i$ are the pre and post correction probability distributions over the classes and \( |\mathcal{C}| \) is the number of samples that get corrected when the chosen concept (or predicate) is replaced by its ground truth. 
In baseline concept-based models, we replace the most misleading concept by its ground truth presence (or absence). In our LogicCBM, since we do not have predicate ground truths, we consider the logic gate of the most misleading logic neuron and compute the selected logic operation on the ground truths of its constituent concepts. This is further described in the Appendix (\cref{subsec:appendix_ccg}) along with how the misleading concept/predicate is estimated. Note that when we make a single correction in a LogicCBM predicate, for fairness of comparison, we make two corrections in other baseline methods. 
\setlength{\intextsep}{8pt}
\setlength{\columnsep}{10pt}
\begin{wraptable}[14]{r}{0\textwidth}
\begin{textsc}
\centering
\footnotesize
\begin{tabular}{|c|c|}
\hline \hline
\textbf{Model} & \textbf{CCG} \\ \hline \hline
Vanilla CBM & 0.2102 \\ 
MLP CBM & 0.117 \\
Boolean CBM & 0.308 \\
LFCBM & 0.2491 \\
Posthoc CBM & 0.293 \\
Sparse CBM & 0.358 \\
\rowcolor{verylightgreen} LCBM (Ours) & \textbf{0.5228} \\ \hline \hline
\end{tabular}
\end{textsc}
\caption{Our CCG metric values on different baseline models on the CUB dataset. Higher the value, the more responsive the model is to corrections.}
\label{tab:ccg_avg_conf_increase}
\end{wraptable}
\cref{tab:ccg_avg_conf_increase} reports the CCG metric for LogicCBM and baseline models (other than VLG-CBMs since we don't have access to their models) on the CUB dataset (more results are in Appendix \cref{sec:archnother_details}). LogicCBMs significantly outperform other methods, showing their utility in potentially high-risk scenarios that require interpretability. 
While these numbers provide a population-level view, we also show some qualitative results at a sample level that examine the pre- and post-correction confidences of these models in \cref{fig:ccg_qualitative}.

\vspace{-4pt}
\section{Conclusion}
\label{sec:conclusion}
\vspace{-4pt}
In this work, we presented LogicCBMs, a new pathway to integrate logic into concept-based learning models. We leverage differentiable fuzzy logic operations integrated into such models to predict a model's classification output through logical compositions over intermediate semantics defined by concepts. To study the performance of our logic-enhanced concept-based models, we perform comprehensive experiments on well-known standard benchmarks as well as on syntheic datasets including a new one we introduce, \textit{CLEVR-Logic}. We also provide a worst-case analysis metric (CCG) that can help support further studies in this area. Logic gates are a way of giving the model more modeling capacity without losing interpretability. Our experiments show that these models improve on multiple model metrics beyond accuracy including better concept alignment, effective interventions and receptivity to corrections. 
We believe that our work provides a new dimension to concept-based learning and can improve a model’s overall performance by giving it the ability to logically express its predictions in terms of semantic symbols.

\vspace{1.5mm}
\noindent \textbf{Acknowledgments.} Deepika SN Vemuri would like
to thank PMRF for the fellowship support. We thank the anonymous reviewers for their helpful feedback in improving the presentation of the paper.
{
    \small
    \bibliographystyle{ieee_fullname}
    \bibliography{main}
}

\renewcommand{\thesection}{A\arabic{section}}
\renewcommand{\thetable}{A\arabic{table}}
\renewcommand{\thefigure}{A\arabic{figure}}
\setcounter{section}{0}

\clearpage
\setcounter{page}{1}
\maketitlesupplementary

\noindent Our code will be made publicly available upon acceptance for further research and reproducibility. In this part of the paper, we provide additional details of our work, including the following information.

\subsection*{Table of Contents}
\addcontentsline{toc}{section}{Table of Contents}
\begin{enumerate}
    \item[A1] {Dataset Details} \dotfill 1
    \item[A2] {Architecture and Implementation Details} \dotfill 1
    \item[A3] {More Results and Ablation Studies} \dotfill 5
    \item[A4] {Limitations and Future Work} \dotfill 5
\end{enumerate}

\section{Dataset Details}
\label{sec:dataset_details}
\begin{itemize}
    \item \textbf{CUB: } The Caltech-UCSD Birds-200-2011 (CUB) dataset \cite{cub_ds} is a fine-grained bird species identification dataset. It consists of 11,788 images of 200 bird categories, 5,994 for training and 5,794 for testing. We use the 312 concepts expert-annotated in the dataset representing bird attributes like beak length and shape, body size, wing color, etc. 
    \item \textbf{AwA2: } The Animals with Attributes (AwA2) dataset \cite{awa2} is commonly used for zero-shot learning (ZSL) and attribute-based classification. It consists of 37322 images (26125 training, 11197 testing) of 50 animal classes, annotated with 85 numeric attribute values for each class and is class-level expert annotated.
    \item \textbf{CIFAR100: } CIFAR100 (Canadian Institute for Advanced Research) is a subset of the Tiny Images dataset \cite{Krizhevsky2009LearningML} comprising 100 classes. It consists of 60000 images, 50000 for training and 10000 for testing. Since this dataset does not have expert annotated concepts, we query an LLM to get 925 concepts. We follow the process outlined by \cite{oikarinen2023labelfree}, where, an initial concept set is generated through queries like "List the most important features for recognizing something as a \{class\}". This is followed by a sequence of filtering steps, to improve the quality of the concept set.
    \item 
    \textbf{SUN Attribute: } The SUN attribute dataset is build on top of the fine-grained SUN (Scene UNderstanding) categorical database \cite{xiao2010sun} used for scene understanding and scene recognition. It contains 14340 images (we use a 80:10:10 train-val-test split) of 539 classes and 102 crowd sourced attributes.
\end{itemize}

\subsection{CLEVR-Logic} \label{sec:clevr_details}
\begin{figure}[t]
\centering
\includegraphics[width=0.9\linewidth]{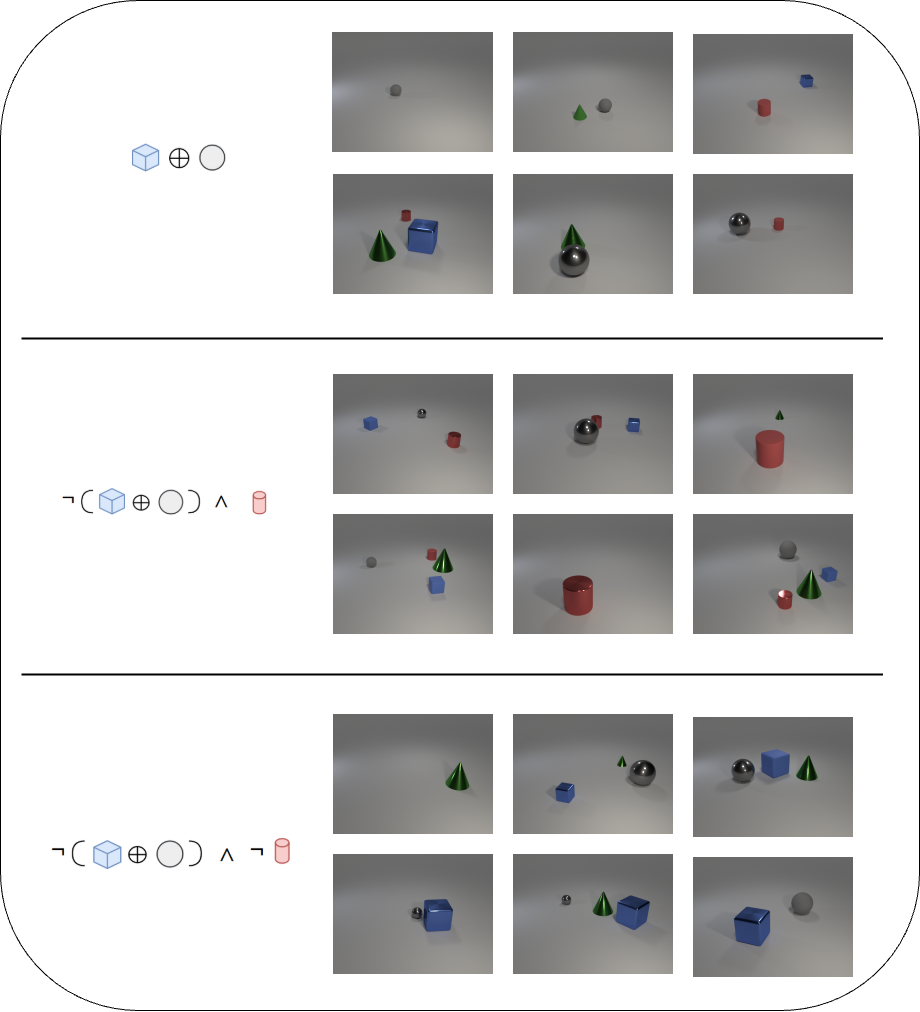}
\caption{More images from our CLEVR-Logic dataset}
\label{fig:clevr_images_appendix}
\end{figure}

As stated in the main paper, we introduce a dataset, CLEVR-Logic, a new variant of CLEVR which we generate to study logical relations among objects in images. We define concepts to be a certain set of \textit{CLEVR} objects (sphere, cone, cube, cylinder) and specify the classes as logical operations among these objects. CLEVR images are then generated per class following the specified class-specific logic. Some example images and the logical relations expected to be learned from them are shown in \cref{fig:clevr_logic_images}. We provide some more images in \cref{fig:clevr_images_appendix}. We found that generating about 20 images per class sufficed for the task considered herein. Our code will be made publicly upon acceptance, and we believe it can be used to generate CLEVR images of greater complexity to capture a wide range of logic operations.

\section{Architecture and Implementation Details}
\label{sec:archnother_details}
\begin{figure}[!t]
    \centering
    \includegraphics[width=250pt]{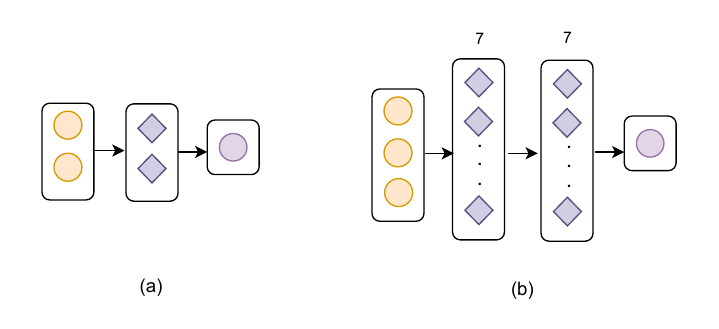}
    \caption{In continuation to \cref{fig:clevr_logic_arch} in the main paper, we provide the logic model architectures used for the \textit{XOR} and \textit{2XOR} synthetic dataset experiments. Yellow circles are concepts, diamonds are logic neurons and purple circles are final outputs.}
    \label{fig:xor_2xor_arch}
\end{figure}

The feature extractor $g$ used in all the Vanilla and Boolean CBM experiments was an Inception V3 \cite{szegedy2016rethinking}, with the concept and classification layers being single linear layers. For all the other baselines, we use their code as is. 
For the CLEVR experiments, we use a ResNet-18 \cite{he2016deep} as the feature extractor. We provide the architectures used for the synthetic dataset experiments in \cref{fig:xor_2xor_arch}. For the XOR and 2XOR experiments, we train the models on the truth tables of both these operations respectively.  All experiments were run on a single NVIDIA GeForce RTX 3090.

\subsection{Additional Details on Metrics}
\label{subsec:appendix_ccg}
\noindent \textbf{Further Details on CCG: } As described in the main paper, CCG is a metric computed to test the worst-case behavior of a model. Concretely, we choose the most misleading concept (or predicate) and turn it off (or on according to its ground truth) and measure the change in the confidence of the model. The most misleading concept (or predicate) is the one that has the most weight difference between the predicted class and ground truth class (\cref{alg:concept_selection}). The CCG computation process is described in \cref{alg:ccg}, where we first choose the most misleading concept (or predicate), replace it with its ground truth (or predicate activation computed on its ground truth concepts) and then pass the concept activations through the classifier again to check whether this leads to a correction in the models prediction. 

\begin{algorithm}
\footnotesize
\captionsetup{font=footnotesize}
\caption{\textsc{Get\_Misleading\_Unit}($y, \hat{y}, W$):}\label{alg:concept_selection}
     \begin{algorithmic}
     \Require Ground truth class $y$,  predicted class distribution $\hat{y}$, classifier $f$'s weights $W$ \\
    \Return $\argmax_i |W[y, i] - W[\hat{y}, i]|$
    \end{algorithmic}
\end{algorithm}

\begin{algorithm}
\footnotesize
\captionsetup{font=footnotesize}
    \caption{Concept Correction Gain Metric}\label{alg:ccg}
    \begin{algorithmic}
        \Require Classifier $f$, penultimate layer activations $a$, classifier $f$'s weights $W$, input image-concepts-label tuple ($x$, $c$, $y$)
        \State $\hat{y} \gets f(a)$
        \State $i \gets  \textsc{Get\_Misleading\_Unit}(y, \hat{y}, W)$
        \If{concept model}
        \State a[i] := c[i] 
        \Else 
        \State a[i] := $\hat{z}$[i](c[a1], c[a2]) \\
        \Comment{Compute predicate on its constituent concept ground truths (a1, a2)}
        \EndIf
        \State $\check{y} \gets f(a)$
    \end{algorithmic}
\end{algorithm}

\begin{table}[h]
\footnotesize
\centering
\begin{tabular}{|c|c|c|c|}
\hline \hline
\textbf{Model} & \textbf{AwA2} & \textbf{CIFAR100} & \textbf{SUN} \\ \hline \hline
Vanilla CBM & 0.377 & 0.078 & 0.252 \\
\hline
LogicCBM & \textbf{0.485} & \textbf{0.410} & \textbf{0.616} \\
\hline \hline
\end{tabular}
\caption{CCG comparison between Vanilla CBM and LogicCBM across different datasets.}
\label{tab:ccg_appendix}
\end{table}

\begin{table*}[h]
\footnotesize
\centering
\begin{tabular}{p{4cm} p{5.5cm} p{5.5cm}}
\toprule & \textbf{Neurosymbolic Programming Frameworks} & \textbf{Our Framework} \\
\midrule
\textbf{Logic specification} 
& Logic rules are pre-specified and fixed, reflecting a symbolic paradigm. 
& Logic gates are learnable components whose configurations emerge through training. \\
\midrule
\textbf{Differentiability vs. learnability} 
& Logic is differentiable but not learnable; only rule probabilities are optimized (e.g., in DeepProbLog). 
& Logic is both differentiable and learnable; gates obtained through gradient-based training. \\
\midrule
\textbf{Symbols / predicates} 
& Neural predicates need not be interpretable; they may correspond to arbitrary latent factors. 
& Concepts are explicitly human-interpretable by design. \\
\midrule
\textbf{Interpretability source} 
& Interpretability arises from the structure of reasoning (e.g., through a domain-specific language). 
& Interpretability arises from human-defined concepts and their learned logical composition. \\
\midrule
\textbf{Typical application domain} 
& Structured reasoning tasks, where constraints can be explicitly encoded as logical rules. 
& Continuous neural domains, such as classification tasks. \\
\bottomrule
\end{tabular}
\caption{Comparison between neurosymbolic programming frameworks and our proposed framework.}
\label{tab:neurosym_comparison}
\end{table*}

\noindent 
We also present some CCG results on the other datasets in \cref{tab:ccg_appendix} and see that LogicCBMs consistently do better than Vanilla CBMs.

\vspace{1.5mm}
\noindent \textbf{Performing corrections with no concept ground truths: } Some baseline methods (Posthoc CBMs \cite{yuksekgonul2023posthoc} and Sparse CBMs \cite{scbm}) in \cref{tab:main_table} do not use concept ground truths in their training process. Although the list of concepts considered overall is provided per dataset, there is no class-level or instance-level concept ground-truth. Since we need concept ground truths to perform a correction for CCG, we propose the process outlined in \cref{alg:no_gt_correction} to perform interventions on these models.
We use a pretrained SentenceTransformer \footnote{https://huggingface.co/sentence-transformers} to get the embeddings of the selected (most misleading) text-concepts and the ground truth text-class, and measure the similarity between these embeddings. The concept is then set to the minimum/maximum concept activation using a threshold on this embedding similarity. We use \textit{cosine similarity} to measure similarity and observe that a threshold of 0.5 worked well in our experiments.

\begin{algorithm}[h]
\footnotesize
\captionsetup{font=footnotesize}
    \caption{Correction without concept ground truths}
    \label{alg:no_gt_correction}
    \begin{algorithmic}
    \Require Concept activation vector $\hat{c}$, chosen misleading unit $ind$, text concepts $concepts_{text}$, text classes $classes_{text}$):
      \If{\texttt{SIMILARITY}($concepts_{text}[ind]$, $classes_{text}$[y]) $>$ $threshold$} 
        \State $\hat{c}[ind] \gets \max \hat{c}$
      \Else
        \State $\hat{c}[ind] \gets \min \hat{c}$
      \EndIf\\
      \Return $\hat{c}$
    \end{algorithmic}
\end{algorithm}

\subsection{Additional Experiments} \label{sec:add_exp_details}
\noindent \textbf{Correlated Concepts for Concept Pairing: }Beyond what was presented in the main paper for concept pairing, we also studied our framework by using correlated concepts for concept pairing.  We describe this strategy in \cref{alg:correlated_concepts}, where $n$ random samples are chosen whose concept activations are obtained using the concept encoder $g$. These activations are aggregated and averaged per concept, of which the top two concepts are then estimated to be correlated. This strategy is compared with our original strategy (where we allowed random pairing of concept activations, which allows us to be efficient) in \cref{tab:concept_pairings}. As shown in the table, this approach does imply some features of the datasets. The accuracies for CUB and AwA2 are nearly the same as the original approach, although there is a drop in performance in CIFAR100. This could indicate that there aren't those many correlated concepts in CIFAR100. This may be an interesting direction to study further.

\begin{table}[h]
\footnotesize
\centering
\begin{tabular}{|c|c|c|c|}
\hline \hline
\textbf{Concept Pairing Strategy} & \textbf{CUB} & \
\textbf{AwA2} & \textbf{CIFAR100} \\ \hline \hline
Original & 81.55 & 90.08 & 68.97 \\
\hline
Correlated & 80.38 & 90.14 & 65.28 \\
\hline \hline
\end{tabular}
\caption{Accuracy results using different concept pairing mechanisms.}
\label{tab:concept_pairings}
\end{table}

\begin{algorithm}
\footnotesize
\captionsetup{font=footnotesize}
\caption{Correlated Concept Selection}
\label{alg:correlated_concepts}
\begin{algorithmic}
\Require Random samples $S = \{x_1, x_2, \dots, x_n\}$, concept encoder $g$, concept activation (act) matrix $A \in \mathbb{R}^{n \times h_2}$.
\State Initialize $C = \mathbf{0} \in \mathbb{R}^m$ 
\For {$i = 1$ to $n$}
    \State $A[i, :] \gets g(x_i)$ 
     \Comment{Get concept act for sample $x_i$.}
    \State $C \gets C + A[i, :]$ 
    \Comment{Accumulate concept acts.}
\EndFor
\State $C \gets C / n$ 
\Comment{Average acts across all samples.}
\State $k_1, k_2 \gets \text{argtop2}(C)$ \\
\Return $(k_1, k_2)$
\end{algorithmic}
\end{algorithm}

\noindent \textbf{Extension to n-ary Predicates: }In all our experiments in the main paper, we used only one logic gate layer. A key advantage of using our framework is that we can stack multiple logic modules to form more expressive predicates, which can further enhance the models' performance. We studied the use of two and three logic modules (specific architectures provided in \cref{tab:multi_layer_archs}) and we report results in \cref{tab:multilayerlogic}. The \#L = 1 row also corresponds to the architectures of the logic gate layers used in \cref{tab:main_table}. These results show that extensions to multiple logic modules, while feasible, require more careful tuning and provide interesting directions for future work.

\begin{table}[h]
\footnotesize
\centering
\begin{tabular}{|c|c|c|c|}
\hline \hline
\textbf{\#L} & \textbf{CUB} & \textbf{AwA2} & \textbf{CIFAR100} \\ \hline \hline
1 & 250 $\times$ 1 & 60 $\times$ 1 & 600 $\times$ 1 \\ \hline
2 & 125 $\times$ 2 & 30 $\times$ 2 & 300 $\times$ 2 \\ \hline
3 & 60 $\times$ 3 & 15 $\times$ 3 & 150 $\times$ 3 \\ \hline \hline
\end{tabular}
\caption{Multi-layer logic architectures. \#L indicates the number of layers.}
\label{tab:multi_layer_archs}
\end{table}

\begin{table}[h]
\begin{sc}
\centering
\begin{footnotesize}
\begin{tabular}{|c|c|c|c|}
\hline \hline
\textbf{\#L} & \textbf{CUB} & \textbf
{AwA2} & \textbf{CIFAR100} \\ \hline \hline
1 & 81.55 & 89.9 & 68.95 \\ \hline
2 & 77.62 & 89.39 & 67.48 \\ \hline
3 & 58.76 & 86.3 & 67.67 \\ \hline \hline
\end{tabular}
\caption{Results of extending LogicCBMs to $n$-ary predicates with multiple logic layers. (Note that the single-module results are our method's performance in \cref{tab:main_table}).}
\label{tab:multilayerlogic}
\end{footnotesize}
\end{sc}
\end{table}

\begin{table}[h]
\footnotesize
\centering
\begin{tabular}{|c|c|c|c|}
\hline \hline
\textbf{Model} & \textbf{CUB} & \textbf{AwA2} & \textbf{CIFAR100} \\ \hline \hline
Fuzzy CBM & 312$\times$200 & 85$\times$50 & 925$\times$100 \\ \hline
MLP CBM & 312$\times$250$\times$200 & 85$\times$60$\times$50 & 925$\times$150$\times$100 \\ \hline
Boolean CBM & 312$\times$200 & 85$\times$50 & 925$\times$100 \\ \hline
LFCBM & 290$\times$200 & 72$\times$50 & 892$\times$100 \\ \hline
Sparse CBM & 370$\times$200 & 85$\times$50 & 944$\times$100 \\ \hline
Posthoc CBM & 312$\times$200 & 85$\times$50 & 440$\times$200 \\ \hline
LogicCBM & 250$\times$200 & 60$\times$50 & 600$\times$100 \\ \hline \hline
\end{tabular}
\caption{The number of parameters in the concept to class layer in all baselines. \#concepts x \#classes.}
\label{tab:cbm_archs}
\end{table}

\begin{figure*}[t]
    \centering
    \includegraphics[width=\linewidth]{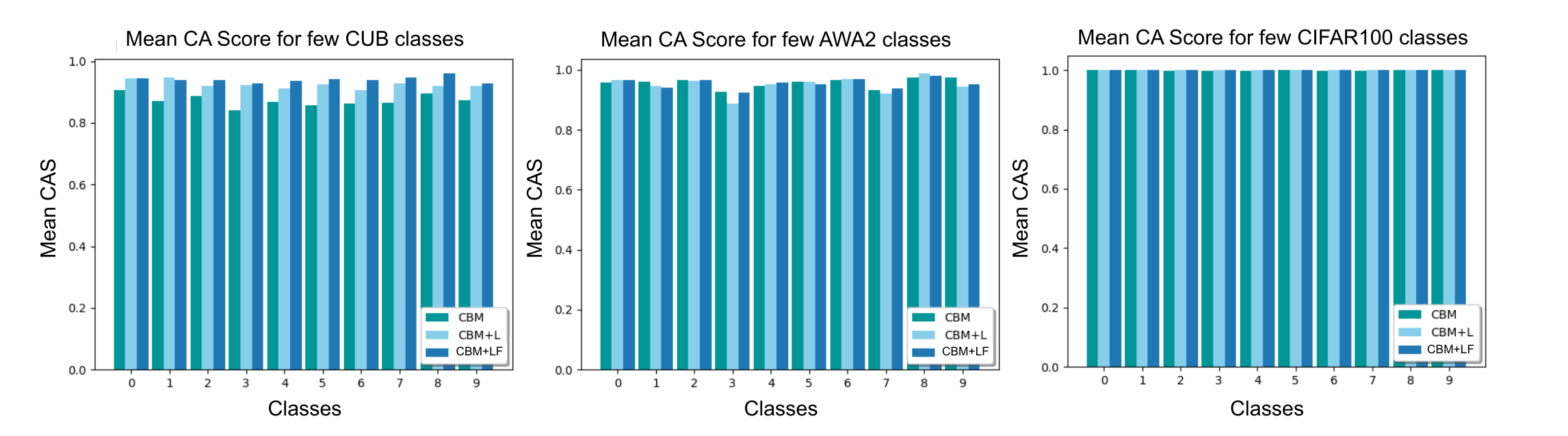}
    \caption{Mean concept alignment scores on 10 random classes (x-axis) of CUB, CIFAR100 and AwA2, on a Vanilla CBM, LogicCBM (CBM+L) and a Vanilla CBM+LF (CBM+LF).}
    \label{fig:cas_plots_appendix}
\end{figure*}

\begin{figure*}[t]
    \centering
    \includegraphics[width=\linewidth]{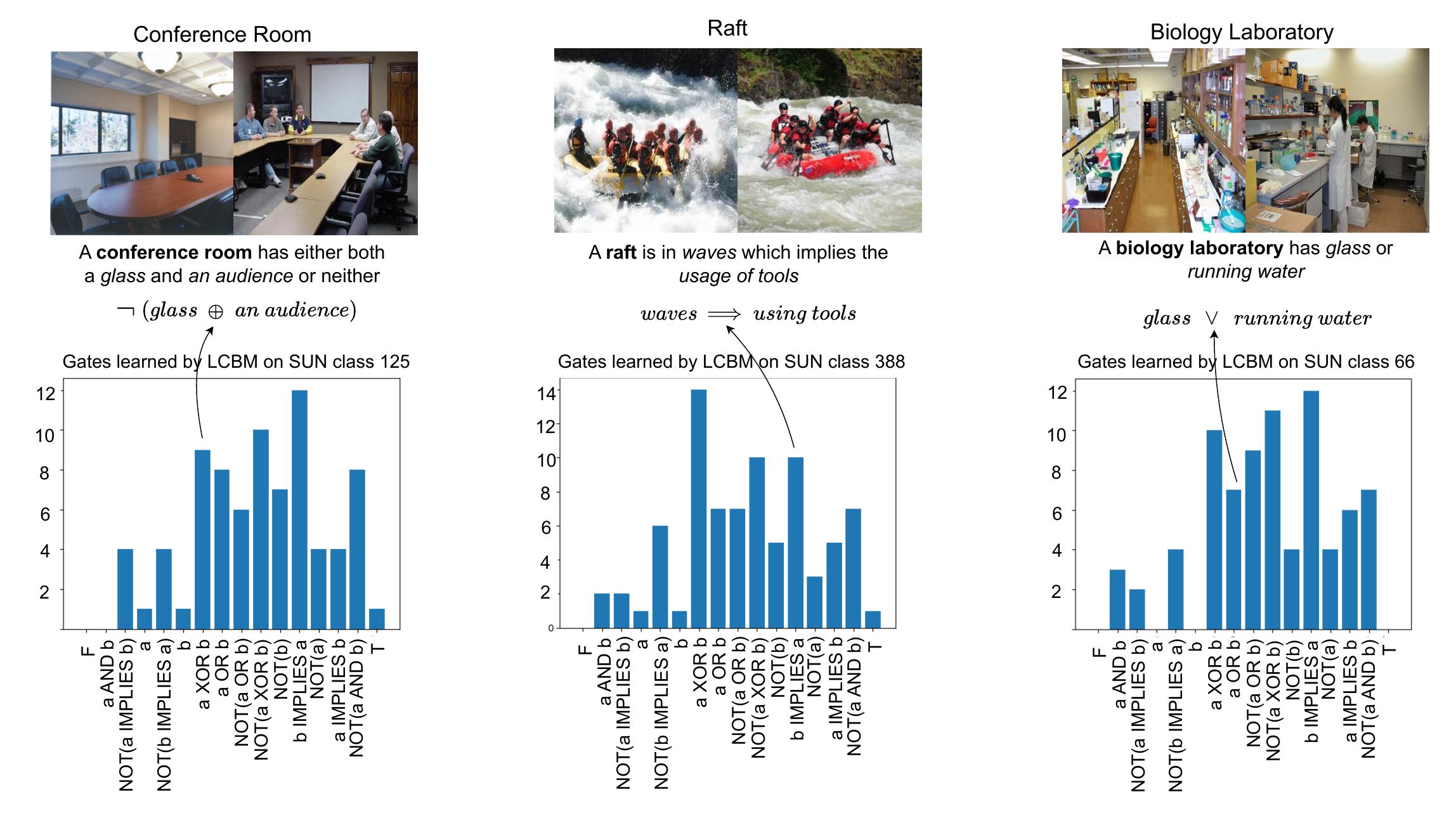}
    \caption{More examples of predicates learned on a LogicCBM for some classes from the SUN attribute dataset, along with the distribution of logic gates learned for those respective classes.}
    \label{fig:sun_result_appendix}
\end{figure*}

\subsection{Hyperparameter Details}
All models were trained with a batch size of 64. The number of parameters in $f$ (concept-class layer) are provided for all baselines in \cref{tab:cbm_archs}. We describe the hyperparameters used for each baseline.
\begin{itemize}
    \item \textit{Vanilla CBMs: }On CUB, the Vanilla CBMs were trained for 40 epochs, with a learning rate 0.001, weight decay of 1e-4 and $\alpha$ as 0.01 using an SGD optimizer. The AwA2 models were trained for 30 epochs, with the other configuration being the same as CUB. The CIFAR100 models were trained for 40 epochs, with a learning rate of 0.001, weight decay of 1e-4 and $\alpha$ as 0.001 using an Adam optimizer. 
    The SUN models were trained for 50 epochs, with a learning rate of 0.01, weight decay of 1e-4 using an SGD optimizer.
    \item \textit{LogicCBMs: }All the LogicCBM models were trained with the same corresponding configuration as the Vanilla CBM models for fairness.
    \item \textit{MLP CBMs: }We trained these models for roughly double the number of epochs as the Vanilla CBMs with all other configurations remaining the same. 
    \item \textit{Boolean CBMs: }All the Boolean CBM models were trained using the same configuration as the corresponding Vanilla CBM models with the only extra step being that the concept activations are thresholded and made binary before sending them to the classifier.
    \item \textit{Label-Free CBMs: }We used the default hyperparameters given in the paper for CUB and CIFAR100. For AwA2 the models were trained for 1000 epochs.
    \item \textit{Posthoc CBMs: }On CUB, the models were trained for 40 epochs, with a learning rate of 0.01. On CIFAR100, the models were trained for 100 epochs, with the same learning rate of 0.001. On AwA2, the models were trained for 100 epochs, with a learning rate of 0.0001.
    \item \textit{Sparse CBMs: }On CUB, the models were trained for 100 epochs, with a learning rate of 3e-4. On CIFAR100, the models were trained for 40 epochs, with the same learning rate of 3e-4. On AwA2, the models were trained for 20 epochs, with a learning rate of 3e-3.
    \item \textit{Logic for Finetuning:} For the results presented in \cref{tab:logic_finetuning}, we use the Vanilla CBM models from row 1 of \cref{tab:main_table}. We finetune the models using our logic module for 15 epochs on CUB, 40 epochs on AwA2 and 10 epochs on CIFAR100.
\end{itemize}

\section{More Results and Ablation Studies}
\label{sec:appendix-moreresults}
\noindent \textbf{Concept Alignment Score. }In Table \ref{tab:cas_scores}, we presented results on the mean concept-alignment scores per benchmark averaged over all their respective classes, in order to measure the average degree of similarity among the concept activations of samples belonging to the same class. Here, we present some mean concept-alignment scores per class in Figure \ref{fig:cas_plots_appendix}. 

\begin{figure}
    \centering
    \includegraphics[width=0.9\linewidth]{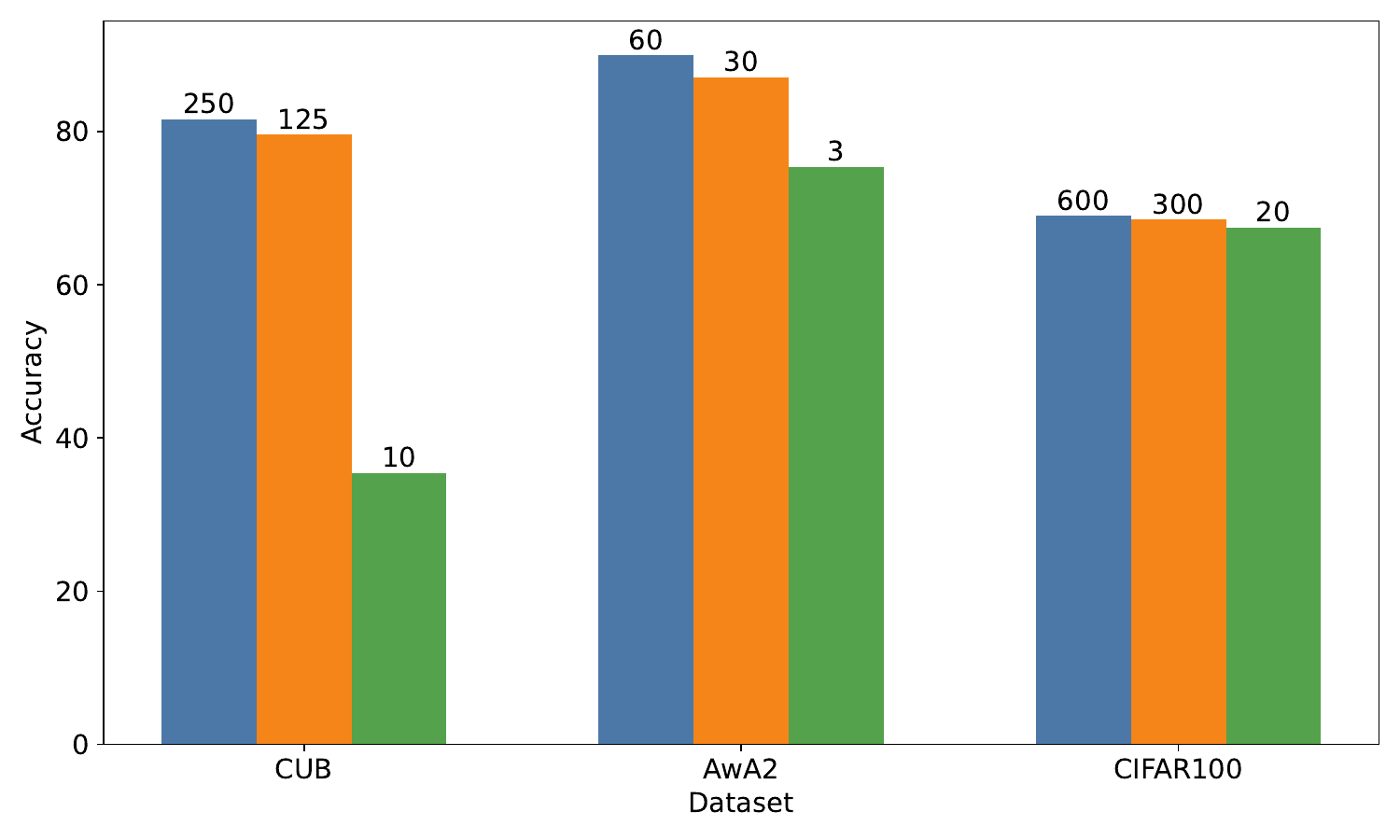}
    \caption{Analysis on the effect of logic layer size on classification accuracy on all datasets.}
    \label{fig:p_analysis}
\end{figure}

\vspace{1mm}
\noindent \textbf{Logic Layer Size. }In \cref{fig:p_analysis}, we present some analysis on the number of logic neurons used in our single logic layer LogicCBM models. As shown in \cref{fig:p_analysis}, we see a consistent drop in performance across datasets when $p$ is decreased. The drop is higher on CUB indicating the complexity of the dataset.

\vspace{1mm}
\noindent \textbf{Choice of $\alpha$ and $\beta$} We perform ablation studies on the two hyperparameters: $\alpha$ (LogicCBM) and $\beta$ (Vanilla CBM+LF). The best LogicCBM and best Vanilla CBM+LF models per dataset are considered, and the values of these hyperparameters are varied among [0.001, 0.01, 0.1, 1]. As seen from the results from Tables \ref{tab:alpha}, \ref{tab:beta}, there is a trend towards a drop in performance as the concept weight is increased. 

\begin{figure}
    \centering
    \includegraphics[width=\linewidth]{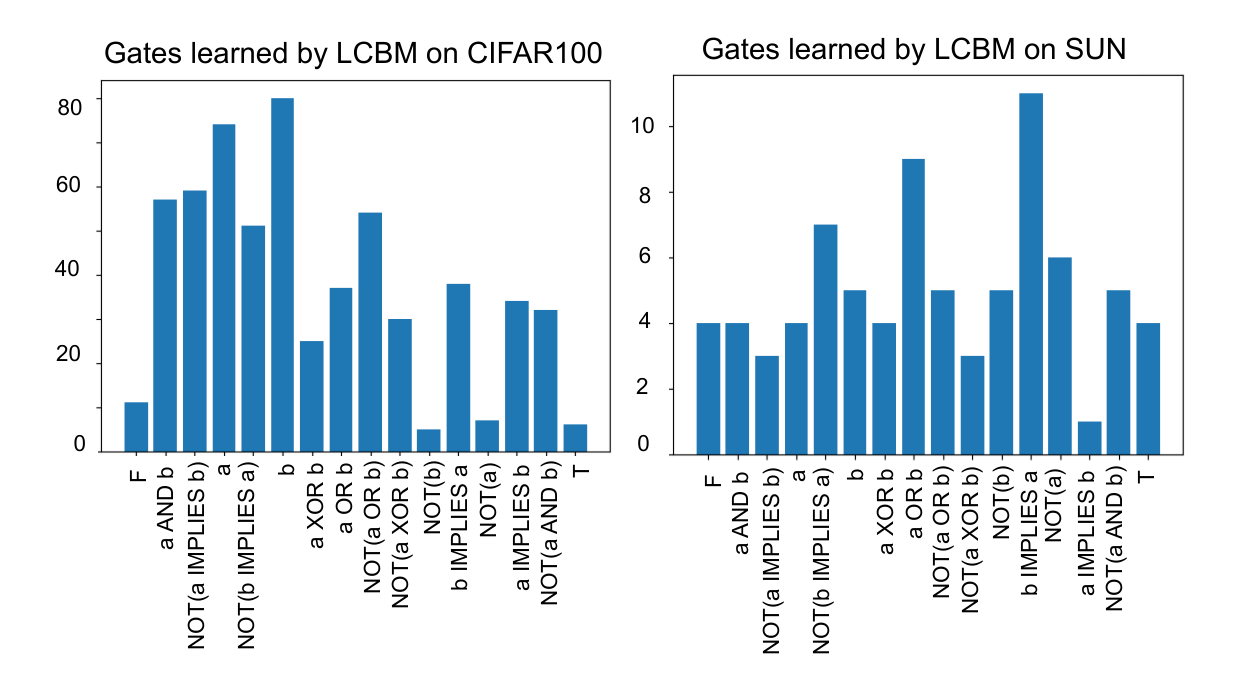}
    \caption{The distribution of logic gates learned by our LogicCBM models on CIFAR100 and SUN.}
    \label{fig:gate_barplot_cifar100_sun}
\end{figure}

\begin{figure*}
    \centering
    \includegraphics[width=450pt]{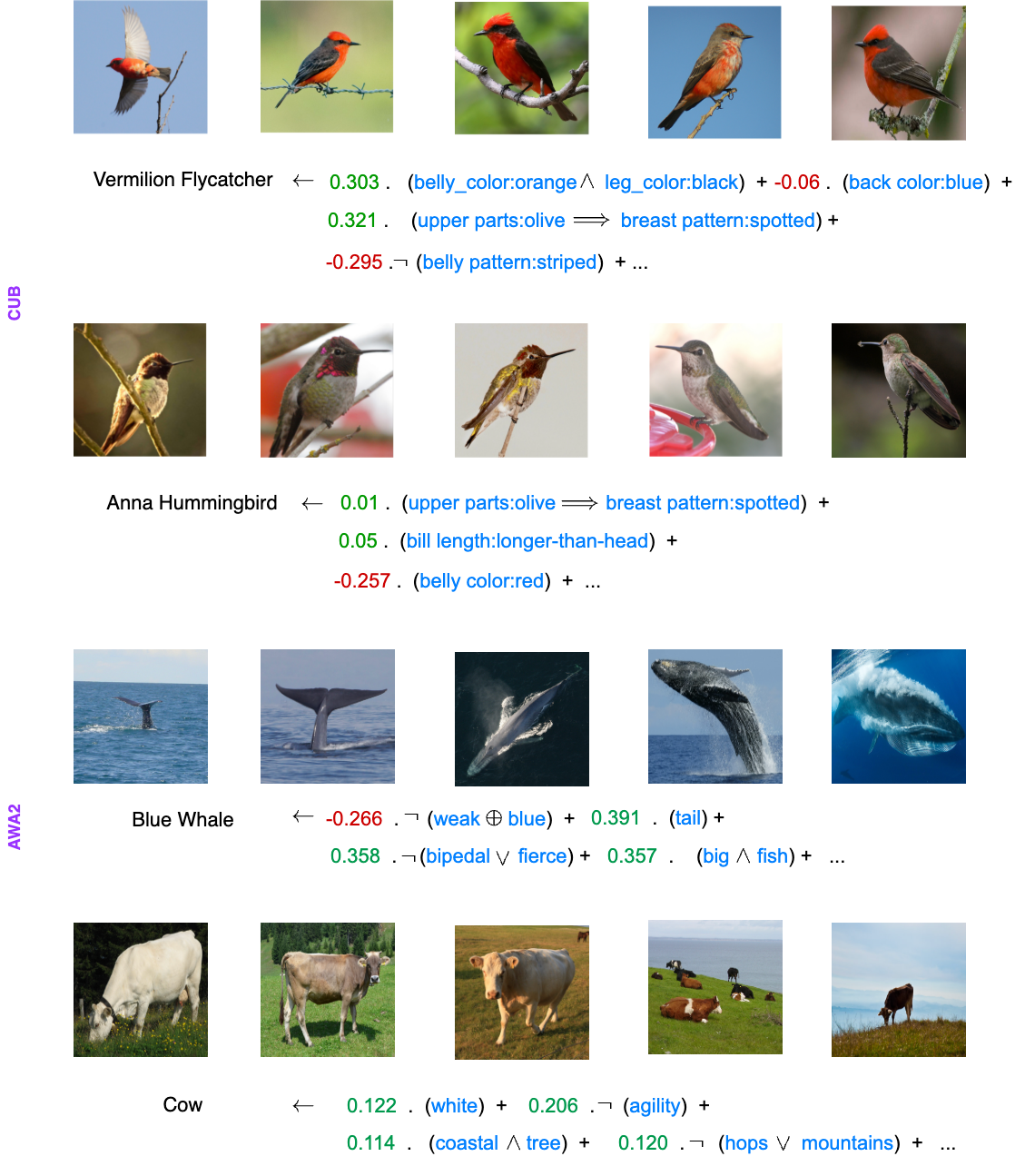}
    \caption{More examples of the class-level logic learned by logic-enhanced CBMs.}
    \label{fig:class_level_logic_appendix}
\end{figure*}

\begin{figure*}
    \centering
    \includegraphics[width=\linewidth]{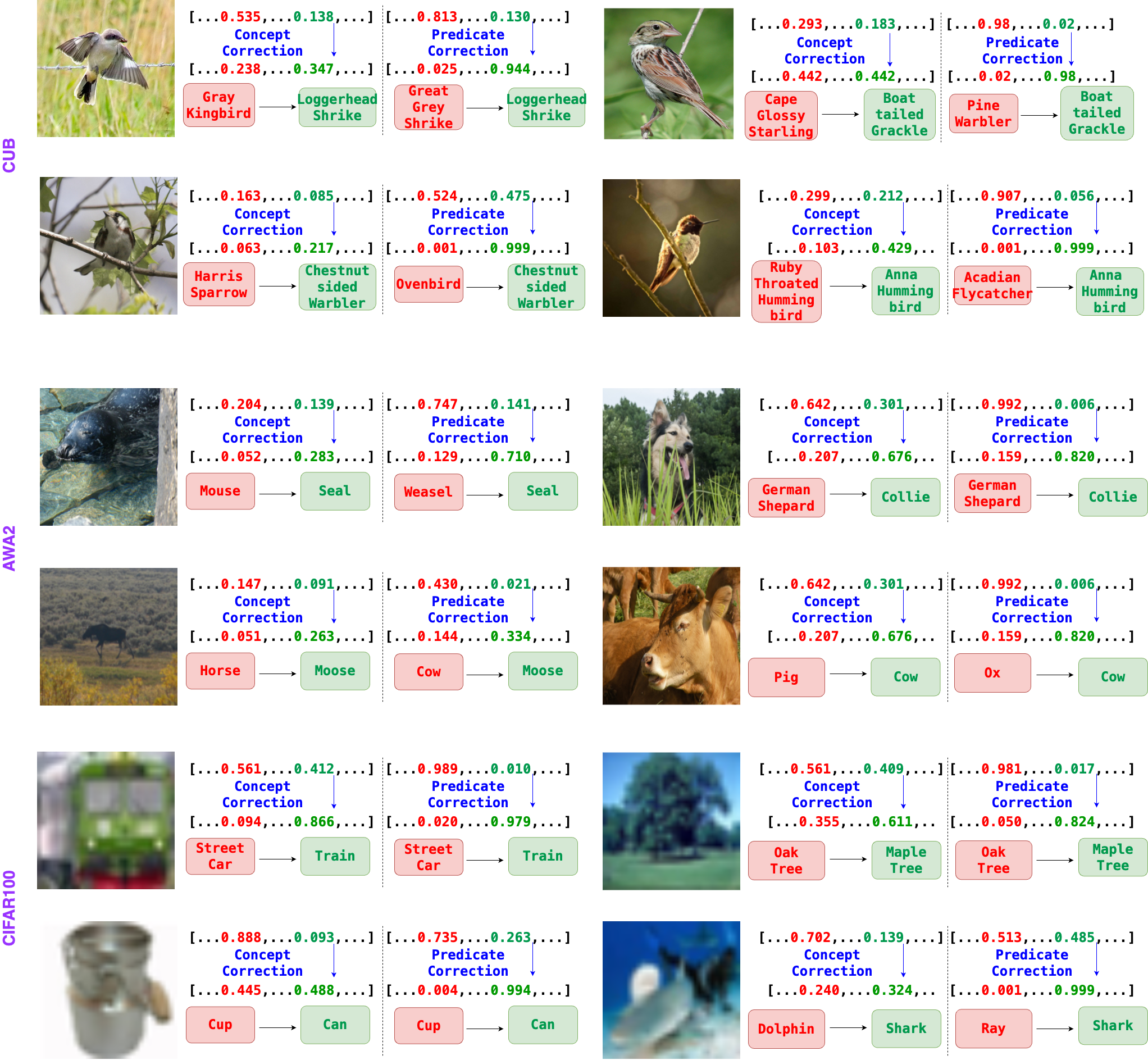}
    \caption{More examples of pre- and post-correction confidences on a CBM and a logic-enhanced CBM across three datasets.}
    \label{fig:ccg_qualitative_appendix}
\end{figure*}

\vspace{1mm}
\noindent \textbf{More qualitative results. } 
We provide the distribution of logic gates learned for the CIFAR100 and SUN attribute dataset in \cref{fig:gate_barplot_cifar100_sun}. More qualitative results on pre- and post-correction confidences on samples from the three datasets are presented in \cref{fig:ccg_qualitative_appendix}. We also show some examples of the class-level logic learned for specific classes in \cref{fig:class_level_logic_appendix}, where the learned logic is presented in the form of Equation \ref{eq:1}. 
Finally, we provide some more example predicates on some classes from the SUN attribute dataset in \cref{fig:sun_result_appendix} along with their corresponding learned class-level logic.
These results show the promise of our framework.

\section{Limitations and Future Work}
The logic module in all our logic-enhanced models (from Table \ref{tab:main_table}) works with a vector of concept activations (scalars), through the presence of an explicit concept layer. Extending it to work with concept embeddings or multi-modal concepts (a vector of vectors), could be a useful future direction. 
We hypothesize that this may cause a drop in some interpretability (as there is no longer a well-defined notion of concept activation in such models), and addressing this challenge would be interesting future work. As shown in our results in Table \ref{tab:concept_pairings}, the use of more intelligent concept pairing mechanisms may yield better outcomes, although this might require further studying. Finally, future work could also further explore the two extensions we propose - logic for finetuning and multi-layered logic (that we do some initial experimentation with in this work). 
We believe that our work herein paves the path for future such efforts in this direction.

\begin{table}[H]
\footnotesize
\centering
\begin{tabular}{|c|c|c|}
\hline \hline
\textbf{Dataset} & \textbf{$\alpha$} & \textbf{CBM+L} \\ \hline \hline
CUB & 0.01 & 81.55 \\ 
 & 0.1 & 47.16 \\ 
 & 1 & 41.9 \\ \hline
AwA2 & 0.001 & 90.05 \\ 
    & 0.1 & 87.64 \\ 
    & 1 & 61.47 \\ \hline  
CIFAR100 & 0.001 & 68.95 \\
    & 0.1 & 60.39 \\
    & 1 & 31.48 \\ \hline \hline
\end{tabular}
\caption{Effect of $\alpha$ on CBM+L on accuracy.}
\label{tab:alpha}
\end{table}

\begin{table}[H]
\footnotesize
\centering
\begin{tabular}{|c|c|c|}
\hline \hline
\textbf{Dataset} & \textbf{Beta} & \textbf{LCBM} \\ \hline \hline
CUB & 0.01 & 81.55 \\ 
 & 0.1 & 65.02 \\
 & 1 & 50 \\ \hline
AwA2 & 0.001 & 90.05  \\
    & 0.1 & 87.18  \\
    & 1 & 65.86 \\ \hline
CIFAR100 & 0.001 & 68.95  \\
    & 0.1 & 56.96  \\
    & 1 & 26.94  \\ \hline \hline
\end{tabular}
\caption{Effect of $\beta$ on LCBM (logic for finetuning) on accuracy.}
\label{tab:beta}
\end{table}

\begin{table*}
    \centering
    \begin{tabular}{rclccccccccc}
        \toprule \hline
        ID    & Operator                    & real-valued   & 00 & 01 & 10 & 11 \\
        \midrule \hline
           0  & False                       & $0$                       & 0     & 0     & 0     & 0     \\
           1  & $A\land B$                  & $A\cdot B$                & 0     & 0     & 0     & 1     \\
           2  & $\neg(A \Rightarrow B)$     & $A-AB$                    & 0     & 0     & 1     & 0     \\
           3  & $A$                         & $A$                       & 0     & 0     & 1     & 1     \\
           4  & $\neg(A \Leftarrow B)$      & $B-AB$                    & 0     & 1     & 0     & 0     \\
           5  & $B$                         & $B$                       & 0     & 1     & 0     & 1     \\
           6  & $A \oplus B$                & $A + B - 2AB$             & 0     & 1     & 1     & 0     \\
           7  & $A \lor B$                  & $A + B - AB$              & 0     & 1     & 1     & 1     \\
           8  & $\neg(A \lor B)$            & $1 - (A+B-AB)$        & 1     & 0     & 0     & 0     \\
           9  & $\neg(A \oplus B)$          & $1 - (A+B-2AB)$       & 1     & 0     & 0     & 1     \\
           10 & $\neg B$                    & $1 - B$                   & 1     & 0     & 1     & 0     \\
           11 & $A \Leftarrow B$            & $1-B+AB$                  & 1     & 0     & 1     & 1     \\
           12 & $\neg A$                    & $1-A$                  & 1     & 1     & 0     & 0     \\
           13 & $A \Rightarrow B$           & $1-A+AB$                  & 1     & 1     & 0     & 1     \\
           14 & $\neg(A \land B)$           & $1 - AB$                  & 1     & 1     & 1     & 0     \\
           15 & True                        & $1$                       & 1     & 1     & 1     & 1     \\
        \bottomrule \hline
    \end{tabular}
    \caption{
        List of real-valued binary logic operations considered in this work \cite{petersen2022deep}.%
        \label{tab:operators}
    }
\end{table*}

\begin{table*}[t]
\centering
\begin{tabular}{|c|c|p{10cm}|}
\hline \hline
\textbf{Dataset} & \textbf{Class} & \textbf{Concepts} \\
\hline \hline
\multirow{3}{*}{\parbox{0.85cm}{\vspace{1.75cm} \hspace{4cm} CUB}} & \multirow{1}{*}{\parbox{3.25cm}{\vspace{0.5cm} Black-footed Albatross}} & 
  back pattern: solid, under tail color: rufous, wing shape: long-wings, belly color: red, wing color: red, upperparts color: brown, breast pattern: multi-colored, upperparts color: rufous, bill shape: cone, tail shape: notched tail, back color: blue \vspace{3pt} \\
& \multirow{1}{*}{\parbox{2.25cm}{\vspace{0.6cm}  American Crow}} & 
  back pattern: solid, wing shape: long-wings, upperparts color: brown, bill shape: cone, tail shape: notched tail, back color: blue, under tail color: grey, wing shape: tapered-wings, belly color: iridescent, wing color: iridescent \vspace{3pt} \\
& \multirow{1}{*}{\parbox{2.25cm}{\vspace{0.5cm} Lazuli Bunting}} & 
  back pattern: solid, under tail color: rufous, throat color: pink, wing shape: long-wings, wing color: red, upper tail color: pink, upperparts color: brown, breast pattern: multi-colored, bill shape: cone \\
\hline
\multirow{3}{*}{\parbox{1.5cm}{\vspace{1.15cm} AwA2}} & \multirow{1}{*}{\parbox{1cm}{\vspace{0.5cm}Raccoon}} & 
  black, white, gray, patches, spots, stripes, furry, small, pads, paws, tail, chewteeth, meatteeth, claws, walks, fast, quadrapedal, active, nocturnal, hibernate, agility \vspace{3pt} \\
& \multirow{1}{*}{\parbox{0.75cm}{\vspace{0.5cm}Cow}} & 
  black, white, brown, patches, spots, furry, toughskin, big, bulbous, hooves, tail, chewteeth, horns, smelly, walks, slow, strong, quadrapedal, active, inactive \vspace{3pt} \\
& \multirow{1}{*}{\parbox{1.25cm}{\vspace{0.25cm} Dolphin}} & 
    white, blue, gray, hairless, toughskin, big, lean, flippers, tail, chewteeth, swims, fast, strong, muscle, active, agility, fish, newworld, oldworld, coastal, ocean, water\\
\hline

\multirow{3}{*}{\parbox{1.5cm}{\vspace{1.15cm} CIFAR100}} 
& \multirow{1}{*}{\parbox{0.75cm}{\vspace{0.25cm} Chair}} & 
  furniture, a person, object, legs to support the seat, an office, a computer, a desk, four legs, a backrest, armrests on either side \vspace{3pt} \\
  & \multirow{1}{*}{\parbox{0.75cm}{\vspace{0.25cm} House}} & 
  windows, building, object, structure, a yard, a chimney, a door, a wall, siding or brick exterior, a garage, roof \vspace{3pt} \\
& \multirow{1}{*}{\parbox{1.25cm}{\vspace{0.25cm} Kangaroo}} & 
  a grassland, short front legs, an animal, a safari, mammal, a long, powerful tail, brown or gray fur, marsupial, long, powerful hind legs, Australia\\
\hline \hline
\end{tabular}
\caption{Some sample classes and a subset of their corresponding concepts for all the real-world datasets.}
\label{fig:appendix-sampleconcepts}
\end{table*}

\end{document}